%% file: main.tex
\pdfoutput=1

\documentclass[11pt]{article}

\usepackage{acl}

\include{headers}

\include{abbreviations}
\include{math_commands}

\title{Randomized Positional Encodings\\Boost Length Generalization of Transformers}

\author{Anian Ruoss$^{*1}$\\\And
    Gr{\'{e}}goire Del\'etang$^{*1}$\\ \And
    Tim Genewein$^{1}$\\ \And
    Jordi Grau-Moya$^{1}$\\ \AND
    R{\'{o}}bert Csord{\'{a}}s$^{\dagger2}$\\ \And
    Mehdi Bennani$^{1}$\\ \And
    Shane Legg$^{1}$\\ \And
    Joel Veness$^{1}$
}

\begin{document}

    \maketitle
    
    \def\thefootnote{*}\footnotetext{Equal contribution. $^1$DeepMind. $^2$The Swiss AI Lab, IDSIA, USI \& SUPSI. $^\dagger$Work performed while the author was at DeepMind. Correspondence to \{anianr, gdelt\}@deepmind.com.}\def\thefootnote{\arabic{footnote}}

    \input{00_abstract}
    \input{01_introduction}
    \input{02_related_work}
    \input{03_method}
    \input{04_results}
    \input{05_conclusion}
    \input{06_limitations}
    \input{07_acknowledgements}

    \bibliography{references}
    
    \clearpage
    \appendix

    \input{08_experimental_details}
    \input{09_additional_results}

\end{document}

%% file: headers.tex
\usepackage{times}
\usepackage{latexsym}

\usepackage[T1]{fontenc}

\usepackage[utf8]{inputenc}

\usepackage{microtype}

\usepackage{amsmath}
\usepackage{amssymb}
\usepackage{booktabs}
\usepackage{graphicx}
\usepackage{multirow}
\usepackage{siunitx}
\usepackage{subcaption}
\usepackage[capitalize]{cleveref}

%% file: abbreviations.tex
\newcommand{\eg}{e.g., }
\newcommand{\ie}{i.e., }

\newcommand{\cf}{cf.\ }

\newcommand{\evenpairs}{\textsc{Even Pairs}}
\newcommand{\modulararithmeticsimple}{\textsc{Modular Arithmetic (Simple)}}
\newcommand{\paritycheck}{\textsc{Parity Check}}
\newcommand{\cyclenavigation}{\textsc{Cycle Navigation}}

\newcommand{\stackmanipulation}{\textsc{Stack Manipulation}}
\newcommand{\reversestring}{\textsc{Reverse String}}
\newcommand{\modulararithmeticbrackets}{\textsc{Modular Arithmetic}}
\newcommand{\solveequation}{\textsc{Solve Equation}}

\newcommand{\duplicatestring}{\textsc{Duplicate String}}
\newcommand{\missingduplicate}{\textsc{Missing Duplicate}}
\newcommand{\oddsfirst}{\textsc{Odds First}}
\newcommand{\binaryaddition}{\textsc{Binary Addition}}
\newcommand{\binarymultiplication}{\textsc{Binary Multiplication}}
\newcommand{\computesqrt}{\textsc{Compute Sqrt}}
\newcommand{\bucketsort}{\textsc{Bucket Sort}}

%% file: math_commands.tex
\usepackage{amsmath,amsfonts,bm}

\def\eqref#1{equation~\ref{#1}}

\def\1{\bm{1}}

\DeclareMathAlphabet{\mathsfit}{\encodingdefault}{\sfdefault}{m}{sl}
\SetMathAlphabet{\mathsfit}{bold}{\encodingdefault}{\sfdefault}{bx}{n}

\def\sN{{\mathbb{N}}}



%% file: 00_abstract.tex
\begin{abstract}
    Transformers have impressive generalization capabilities on tasks with a fixed context length.
    However, they fail to generalize to sequences of arbitrary length, even for seemingly simple tasks such as duplicating a string.
    Moreover, simply training on longer sequences is inefficient due to the quadratic computation complexity of the global attention mechanism.
    In this work, we demonstrate that this failure mode is linked to positional encodings being out-of-distribution for longer sequences (even for relative encodings) and introduce a novel family of positional encodings that can overcome this problem.
    Concretely, our randomized positional encoding scheme simulates the positions of longer sequences and randomly selects an ordered subset to fit the sequence's length.
    Our large-scale empirical evaluation of \num{6000} models across $15$ algorithmic reasoning tasks shows that our method allows Transformers to generalize to sequences of unseen length (increasing test accuracy by $12.0\%$ on average).
\end{abstract}

%% file: 01_introduction.tex
\section{Introduction}
\label{sec:introduction}

Transformers are emerging as the new workhorse of machine learning as they underpin many recent breakthroughs, including sequence-to-sequence modeling~\cite{vaswani2017attention}, image recognition~\cite{dosovitskiy2021image}, and multi-task learning~\cite{reed2022generalist}.
However, recent work~\citep{deletang2023neural} demonstrated that Transformers fail to generalize to longer sequences on seemingly simple tasks such as binary addition.
Thus, while certain problems can be solved without length generalization, algorithmic reasoning generally requires this ability, similar to many real-world settings such as online or continual learning.

\begin{figure}
    \begin{center}
        \includegraphics[width=0.5\textwidth]{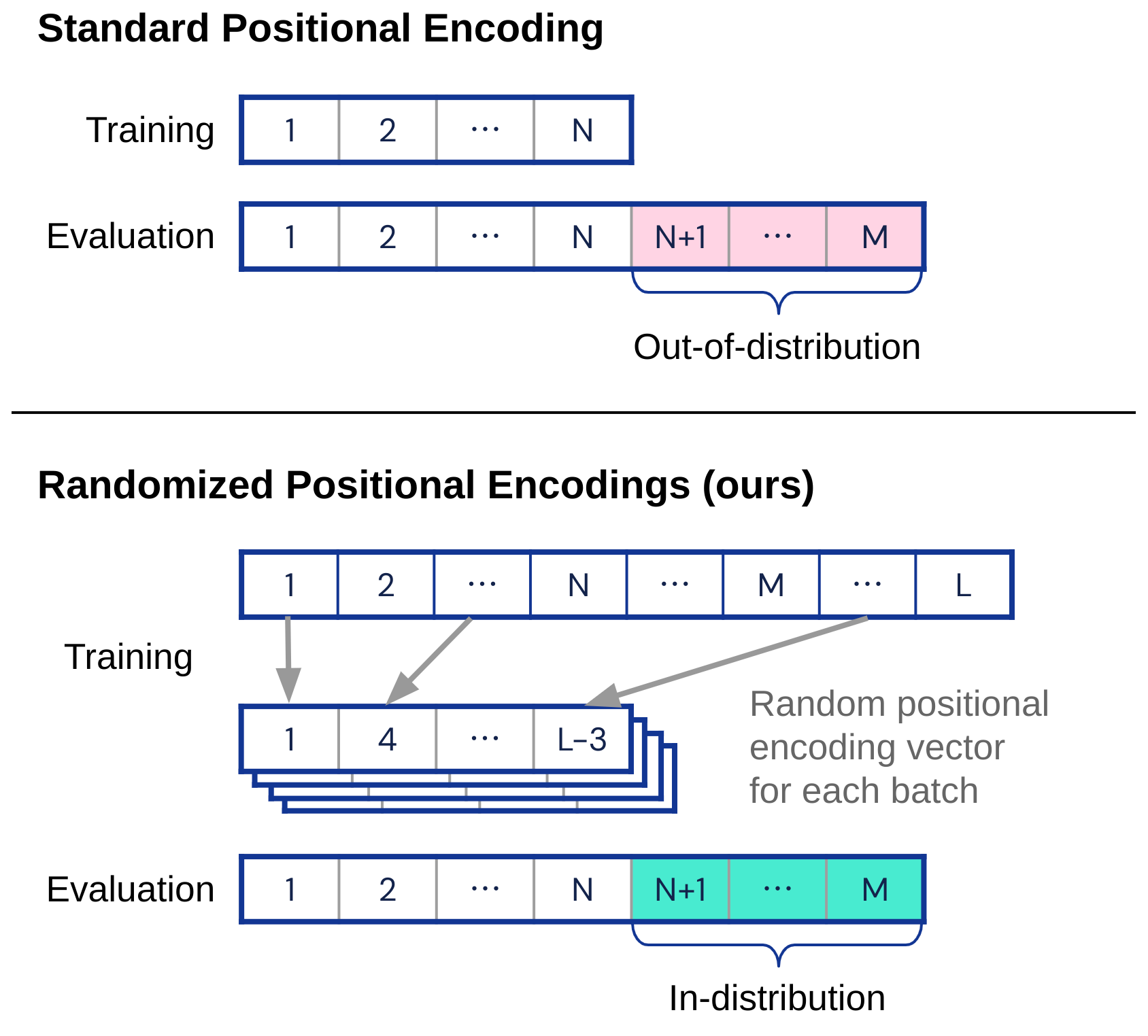}
    \end{center}
    \caption{
        \textbf{Test-time evaluation with longer inputs.} 
        The standard positional encoding vector has values larger than those observed during training. Our approach avoids this problem by assigning a random (ordered) positional encoding vector using the full range of possible test positions to each training example.
    }
    \label{fig:overview}
\end{figure}

While the Transformer's attention mechanism can recognize complex relationships amongst tokens in the input sequence, it is limited by its lack of positional awareness.
Thus, the input sequence is generally augmented with \emph{positional encodings} to inject position information into the computation.
However, current approaches only consider positions up to the maximum training sequence length $N$, and thus all the positions $N+1, \ldots, M$ for test sequences of length up to $M$ will appear out-of-distribution during evaluation (top of \cref{fig:overview}).

\paragraph{This work}

We introduce a novel family of \emph{randomized positional encodings}, which significantly improves Transformers' length generalization capabilities on algorithmic reasoning tasks.
Our approach is compatible with any existing positional encoding scheme and augments the existing methods by subsampling an ordered set of positions from a much larger range of positions than those observed during training or evaluation (\ie up to $L \gg M$; bottom of \cref{fig:overview}).
Thus, over the course of training, the Transformer will learn to handle very large positional encodings and, therefore no longer encounter out-of-distribution inputs during evaluation.
Importantly, our method leaves in-domain generalization performance unaffected and is also significantly more efficient than the naive approach of simply training the Transformer on longer sequences.
Our main contributions are:

\begin{itemize}
    \item A novel family of positional encoding schemes that significantly improves the length generalization capabilities of Transformers, while leaving their in-domain generalization performance unaffected.
    \item A large-scale empirical evaluation on a wide range of algorithmic reasoning tasks showing the superiority of our method over prior work (an increase of the test accuracy 
    by 12.0\% on average and up to 43.5\% on certain tasks).
    \item An open-source implementation of our method, available at \url{https://github.com/deepmind/randomized_positional_encodings}.
\end{itemize}

%% file: 02_related_work.tex
\section{Related Work}
\label{sec:related-work}

Our work is most closely related to the growing line of research on Transformers' positional encodings.
The first approaches simply added a transformation of the tokens' positions, \eg scaled sinusoids~\citep{vaswani2017attention} or learned embeddings~\citep{gehring2017convolutional}, to the embeddings of the input sequence.
\citet{dai2019transformer} subsequently showed that computing the attention (at every layer) using the relative distances between the key and query vectors improves the modeling of long-term (inter-context) dependencies.
Similarly, \citet{su2021roformer} proposed to inject position information by rotating the key-query products according to their relative distances.
Finally, \citet{press2022train} improved the length generalization on natural language processing tasks by adding a constant bias to each key-query attention score (proportional to their distance).
However, as our experiments in \cref{sec:experimental-evaluation} will show, these approaches fail at length generalization on algorithmic reasoning tasks, which is precisely the goal of our work.

A concurrent work developed randomized learned positional encodings~\citep{li2022systematic}, which are a special case of our family of randomized positional encodings.
We also note that the necessity of feature and position randomization for length generalization has been discussed in the context of graph neural networks, which subsume Transformers~\citep{ibarz2022generalist, sato2021random}.
Finally, \citet{liu2020learning} proposed to model the position information as a continuous dynamical system in an effort to handle sequences longer than those seen during training time.

Our work is also related to the research area on improving the systematic (length) generalization capabilities of Transformers~\cite{ontanon2022making}, which includes approaches investigating embedding scaling or early stopping~\citep{csordas2021devil}, adaptive computation time~\citep{dehghani2019universal}, geometric attention with directional positional encodings and gating~\citep{csordas2022neural}, and hierarchical reinforcement learning~\citep{liu2020compositional}.
Such length generalization studies are often conducted in the context of formal language theory, and we evaluate our method on the recent benchmark by \citet{deletang2023neural}, which unifies a large body of work on Transformers' capability to recognize formal languages~\citep{ackerman2006survey, bhattamishra2020ability, ebrahimi2020how, hahn2020theoretical, hao2022formal, merrill2019sequential, merrill2022logprecision}.

%% file: 03_method.tex
\section{Randomized Positional Encodings}
\label{sec:method}

Unlike RNNs~\citep{elman1990finding}, which are unrolled over tokens one step at a time, Transformers process large chunks of the input sequence in parallel via global attention~\citep{vaswani2017attention}.
As a result, Transformers do not need to ``remember'' previous tokens, but they do have to break the permutation-invariance of the attention mechanism.
To that end, the embeddings of the input sequence are generally augmented with positional encodings.
For example, the vanilla Transformer adds the following positional encodings to the embedded input sequence before passing it to the attention layers:
\begin{align}
    \mathrm{PE}(\mathrm{pos}, 2i) &= \sin\left(\frac{\mathrm{pos}}{10000^{\frac{2i}{d_{\mathrm{model}}}}}\right), \label{eq:sin-pos-enc} \\
    \mathrm{PE}(\mathrm{pos}, 2i + 1)  &= \cos\left(\frac{\mathrm{pos}}{10000^{\frac{2i}{d_{\mathrm{model}}}}}\right), \label{eq:cos-pos-enc}
\end{align}
where $\mathrm{pos}$ is the token's position in the sequence, $d_{\mathrm{model}} \in \sN$ is the dimension of the input embedding, and  $i \in \left\{1, 2, \ldots, d_{\mathrm{model}} / 2\right\}$.

While positional encodings generally succeed at inducing the required positional information for sequences of fixed length, they are one of the main failure modes preventing length generalization.
Concretely, for a Transformer with standard positional encodings trained on a curriculum of sequences of maximum length $N$, test sequences of length $M > N$ will shift the distribution of the resultant positional encodings away from those seen in training, with the shift getting increasingly large as $M$ grows.
To address this, we propose a randomized encoding scheme, which relies only on order information, and can be expected to generalize up to sequences of length $M$, where $N < M \leq L$, with a configurable hyperparameter $L$.

\paragraph{Randomized positional encodings}

We assume that each training step will perform a step of loss minimization on a batch of data of fixed size.
Let $\mathcal{U}(S)$ denote the discrete uniform distribution over set $S$, and let $P_k := \{ S \subseteq \{ 1, \dots, L \} \mid |S| = k\}$.
For each training step, we first sample a random length $n \sim \mathcal{U}(\{ 1, \dots, N \})$ (following \citealp{deletang2023neural}) and then a random set of indices $I \sim \mathcal{U}(P_n)$.
We then sort $I$ in ascending order, such that $I = \{i_1, \dots, i_n\}$ for $i_1 < i_2 < \dots < i_n$, noting that $I$ is sampled without replacement.
Finally, we compute our randomized positional encoding for token $1 \leq j \leq N$ as $\mathrm{RPE}(j, \cdot) := \mathrm{PE}(i_j, \cdot)$.
At test time, when processing a sequence of length $M > N$, we use the same procedure but for all token positions $1 \leq j \leq M$.
The intuition behind our method is to preserve the known good properties of relative encoding but in a way that is independent of the maximum training length $N$ and thus allows generalization to longer sequences at test time.

When applying our randomized positional encoding scheme, we subsample the extended positions only once per batch and not individually for every sequence.
For the $\sin / \cos$~\citep{vaswani2017attention}, learned~\citep{gehring2017convolutional}, and RoPE encodings~\citep{su2021roformer}, we apply our method as described above, \ie we directly replace the original token positions with their sampled counterpart.
For the relative encoding~\cite{dai2019transformer}, we compute the relative distances between the sampled positions instead of the original positions.
Finally, for ALiBi~\cite{press2022train}, we sample the bias values from the set of extended positions.

As a consequence, our tokens' positional encodings are no longer directly related to their exact position (the encodings even change during training as they are resampled at every step).
However, since we maintain the order of the encodings, the Transformer can still learn to extract the relevant positional information from the subsampled encodings.
Indeed, we validate the necessity of ordering the sampled positions in our ablation study in \cref{sec:additional-results:ablation}.
Thus, the success of our encoding scheme offers an interesting insight into the inductive biases of the Transformer architecture.

As we will show in \cref{sec:experimental-evaluation}, our randomized encodings trained only on lengths up to $N$ perform the same on sequences of length $M$ as prior approaches trained on lengths up to $M$.
Therefore, our method demonstrates that Transformers can be efficiently trained on short sequences as long as (i) the longer sequences share the same structure and (ii) the longer positions are observed during training.
Moreover, as the running time of global attention is $\mathcal{O}(\ell^2)$ for sequence length $\ell$, our encoding scheme is significantly faster than directly training a model on long sequences.
Furthermore, we also note that our randomized positional encoding scheme significantly boosts length generalization while leaving the in-domain generalization performance largely unaffected (see \cref{fig:scores-encodings}).

The main limitation of our approach is that the maximum test sequence length $M$ has to be known in advance to choose $L \gg M$.
However, our method is compatible with a wide range of values for $L$ (see \cref{sec:additional-results:ablation}), and we note that this is a much weaker assumption than that required for the naive approach of simply training on longer sequences.
However, note that if $L$ is chosen to be much larger than $N$ or $M$, it is theoretically unlikely for the model to encounter enough unique indices during training, likely leading to poor performance (both in- and out-of-distribution).

%% file: 04_results.tex
\section{Experimental Evaluation}
\label{sec:experimental-evaluation}

\paragraph{Problem setup}

\begin{table*}
    \caption{
        Accuracy (in percentage) averaged over all test lengths and maximized over 10 random seeds and 3 learning rates.
        The random accuracy is 50\%, except for \modulararithmeticsimple{}, \cyclenavigation{}, \bucketsort{}, and \modulararithmeticbrackets{}, where it is 20\%.
        Our randomized method increases the test accuracy by 12.0\% on average.
        The randomized learned encodings (denoted with $\star$) are equivalent to label-based encodings~\citep{li2022systematic}.
        $\dagger$ denotes permutation-invariant tasks, which can be solved without positional information.
    }
    \begin{center}
        \resizebox{\linewidth}{!}{\input{tables/scores}}
    \end{center}
    \label{tab:scores}
\end{table*}

We closely follow the experiment setup of \citet{deletang2023neural} and evaluate our method on a wide range of algorithmic reasoning tasks such as modular arithmetic, reversing/duplicating a string, binary addition/multiplication, and bucket sort.
The tasks are derived from formal language recognition and thus grouped according to the Chomsky hierarchy~\citep{chomsky1956three}, which partitions languages into regular (R), context-free, context-sensitive (CS), and recursively enumerable.
Regular tasks can be solved by a finite-state automaton (FSA), deterministic context-free (DCF) tasks can be solved by an FSA with access to a deterministic stack, and CS tasks can be solved by an FSA with access to a bounded tape.
Note that the relation to the Chomsky hierarchy is largely irrelevant for our work and only included for completeness.
We evaluate our method on \citet{deletang2023neural}'s benchmark as it is currently out of reach for Transformers and clearly demonstrates their failure to generalize on algorithmic reasoning tasks.
We refer interested readers to the original paper for more details.

We consider the encoder-only model of the original seq-to-seq Transformer~\citep{vaswani2017attention}, as used in popular pre-trained language models such as BERT~\citep{devlin2019bert} or Gopher~\citep{rae2021scaling}.
Thus, for tasks that require a multi-token output sequence $\bm{y}$ (\eg duplicating a string), we pad the input sequence with $|\bm{y}|$ empty tokens and compute the entire Transformer output from the padded sequence (\ie we do not use autoregressive sampling).
We train the model on sequences of length sampled uniformly from $\mathcal{U}(1, N)$, with $N = 40$, and evaluate it on sequences of length $\left\{N + 1, \ldots, M\right\}$, with $M = 500$.
We set the maximum position $L = 2048$ (and visualize the impact of other values on the performance in \cref{sec:additional-results:ablation}).
We report the accuracy averaged over all unseen sequence lengths, \ie $N + 1, \ldots, M$, for the best-performing model out of 10 different parameter initialization seeds and three learning rates {\num{1e-4}, \num{3e-4}, \num{5e-4}}.
We use the same hyperparameters as~\citet{deletang2023neural} and provide the full experiment setup in \cref{sec:experimental-details}.
We make our code publicly available at \url{https://github.com/deepmind/randomized_positional_encodings}.

\paragraph{Comparison to prior work}

We compare our method to a wide range of positional encodings: none, $\sin / \cos$~\citep{vaswani2017attention}, relative~\cite{dai2019transformer}, ALiBi~\cite{press2022train}, RoPE~\citep{su2021roformer}, learned~\citep{gehring2017convolutional}, and label-based~\citep{li2022systematic}.
Note that the label encodings proposed by \citet{li2022systematic} are equivalent to randomized learned positional encodings and thus subsumed by our method.
We instantiate our randomized positional encoding scheme with all the above encodings and show the average test accuracy in \cref{tab:scores} (with performance curves over test lengths in \cref{sec:additional-results:comparison}).
We observe that our randomized versions significantly increase the test accuracy across most tasks (by 12.0\% on average and up to 43.5\%).
In particular, the randomized relative encoding solves tasks that were previously out of reach for prior work (\eg \reversestring{} or \missingduplicate{}).

\paragraph{Efficiency comparison}

\begin{figure}
    \begin{center}
        \includegraphics[width=\columnwidth]{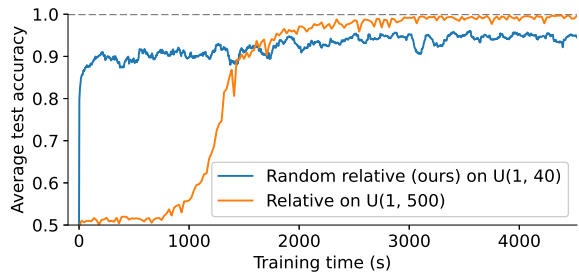}
    \end{center}
    \caption{
        Average accuracy over unseen test lengths on the \missingduplicate{} task over training time (seconds) for two models: (i) our randomized relative positional encoding with a maximum training sequence length of 40, and (ii) the classical relative positional encoding but with a maximum training length of 500.
    }
    \label{fig:efficiency}
\end{figure}

We now show that our method allows us to train a model on short sequences and obtain a test accuracy above 90\%, roughly 35.4 times faster than the naive approach of training a model on longer sequences.
To that end, we train the randomized relative encodings on sequences up to length 40 and the classical relative positional encoding~\cite{dai2019transformer} on sequences up to length 500 and show the test accuracy (averaged over lengths 41 to 500) in \cref{fig:efficiency} over training time (in seconds).
Our model obtains a strong test accuracy significantly faster due to the quadratic cost (in terms of sequence length) of global attention, which means that our model trains at 168.4 steps per second compared to 22.1 steps per second for the naive approach (on a NVIDIA V100 GPU).

%% file: tables/scores.tex
\begin{tabular}{@{}llcrrrrrrcrrrrr@{}}
    \toprule
    & & && & & & & && \multicolumn{5}{c}{\textbf{Randomized (Ours)}} \\
    \cmidrule{11-15}
    \textbf{Level} & \textbf{Task} && \textbf{None} & $\sin/\cos$ & \textbf{Relative} & \textbf{ALiBi} & \textbf{RoPE} & \textbf{Learned} && \textbf{$\sin/\cos$} & \textbf{Relative} & \textbf{ALiBi} & \textbf{RoPE} & \textbf{Learned$^\bigstar$} \\
    \midrule
    \multirow{4}{*}{R}
    & \evenpairs{} && 50.4 & 50.9 & 96.4 & 67.3 & 51.0 & 50.7 && \textbf{100.0} & \textbf{100.0} & 81.5 & \textbf{100.0} & 97.5 \\
    & \modulararithmeticsimple{} && 20.1 & 20.5 & 21.8 & 24.2 & 21.6 & 20.2 && 25.7 & \textbf{28.1} & 21.2 & 25.5 & 21.1 \\
    & \paritycheck{}$^\dagger$ && 51.9 & 50.5 & 51.8 & 51.7 & 51.3 & 50.3 && \textbf{52.6} & 52.2 & 50.3 & 52.3 & \textbf{52.6} \\
    & \cyclenavigation{}$^\dagger$ && 61.9 & 26.3 & 23.0 & 37.6 & 23.6 & 24.2 && 59.0 & 58.8 & 29.8 & \textbf{73.6} & 49.7 \\
    \\
    \multirow{4}{*}{DCF}
    & \stackmanipulation{} && 50.3 & 50.1 & 53.6 & 57.5 & 51.2 & 49.2 && 72.8 & \textbf{77.9} & 70.6 & 68.2 & 69.1 \\
    & \reversestring{} && 52.8 & 50.6 & 58.3 & 62.3 & 51.9 & 50.7 && 75.6 & \textbf{95.1} & 77.1 & 69.9 & 52.9 \\
    & \modulararithmeticbrackets{} && 31.0 & 28.3 & 30.3 & 32.5 & 25.1 & 25.1 && 33.8 & \textbf{34.9} & 31.3 & 32.7 & 31.9 \\
    & \solveequation{} && 20.1 & 21.0 & 23.0 & 25.7 & 23.1 & 20.4 && 24.5 & \textbf{28.1} & 22.0 & 24.5 & 22.1 \\
    \\
    \multirow{7}{*}{CS}
    & \duplicatestring{} && 52.8 & 50.7 & 51.7 & 51.3 & 50.9 & 50.8 && 72.4 & \textbf{75.1} & 68.9 & 68.9 & 53.0 \\
    & \missingduplicate{} && 52.5 & 51.3 & 54.0 & 54.3 & 56.5 & 51.0 && 52.5 & \textbf{100.0} & 79.7 & 88.7 & 52.7 \\
    & \oddsfirst{} && 52.8 & 51.6 & 52.7 & 51.4 & 51.3 & 50.6 && 65.9 & \textbf{69.3} & 64.7 & 65.6 & 52.7 \\
    & \binaryaddition{} && 50.1 & 49.8 & 54.3 & 51.4 & 50.4 & 49.8 && 64.4 & \textbf{64.5} & 56.2 & 60.2 & 61.7 \\
    & \binarymultiplication{} && 49.9 & 50.1 & \textbf{52.2} & 51.0 & 50.2 & 49.6 && 52.1 & 50.1 & 50.5 & 51.7 & 51.9 \\
    & \computesqrt{} && 50.2 & 50.1 & 52.4 & 50.9 & 50.5 & 50.2 && 52.5 & \textbf{53.3} & 51.2 & 52.3 & 52.0 \\
    & \bucketsort{}$^\dagger$ && 23.7 & 30.1 & 91.9 & 38.8 & 30.6 & 25.9 && \textbf{100.0} & \textbf{100.0} & 99.6 & 99.6 & 99.5 \\
    \bottomrule
\end{tabular}

%% file: 05_conclusion.tex
\section{Conclusion}
\label{sec:conclusion}

We introduced a novel family of positional encodings that significantly improves the length generalization capabilities of Transformers.
Our positional encodings are based on the insight that conventional positional encodings will be out-of-distribution when increasing the sequence length.
Thus, to overcome this issue, we randomly sample our encodings from a wider range than the lengths seen at test time while keeping the order.
Our large-scale empirical evaluation demonstrates that our method significantly outperforms prior work in terms of length generalization while offering superior computational performance over the naive approach of training the model on longer sequences.

%% file: 06_limitations.tex
\section*{Limitations}
\label{sec:limitations}

While our work shows promising results in improving the generalization capabilities of Transformers to sequences of arbitrary length, some limitations must be considered.
First, our evaluation is confined to synthetic algorithmic reasoning tasks, which may not fully capture the complexity and diversity of natural language.
We focused on synthetic datasets since they showed clear and somewhat surprising limitations of Transformer architectures~\citep{deletang2023neural}.
However, the generalizability of our approach to other tasks and domains remains an open question, and additional research, such as evaluation on SCAN~\citep{lake2018generalization}, CFQ~\citep{keysers2020measuring}, COGS~\citep{kim2020cogs}, or the Long Range Arena~\citep{tay2021long}, is necessary to understand its potential in real-world applications.
Second, our approach introduces a new hyperparameter -- the maximum sequence position $L$.
Although our experiments in \cref{sec:additional-results:ablation} show that our method's performance is largely unaffected by the precise value of $L$, practitioners may still have to tune the parameter depending on their specific problem domains.
Third, we only isolate and ameliorate one failure mode of Transformer length generalization on synthetic datasets.
However, there are other factors contributing to poor length generalization, such as attention becoming less peaked for longer sequences~\citep{chiang2022overcoming}.
Overall, we believe that our study's limitations offer several interesting directions for future research.

%% file: 07_acknowledgements.tex
\section*{Acknowledgements}
\label{sec:acknowledgements}

We thank Chris Cundy, Elliot Catt, Kevin Li, Laurent Orseau, Marcus Hutter, Petar Veli\v{c}kovi\'{c}, Vincent Dutordoir, and the anonymous reviewers for their helpful feedback.

%% file: 08_experimental_details.tex
\section{Experimental Details}
\label{sec:experimental-details}

We use the experiment suite proposed by \citet{deletang2023neural}, which consists of 15 algorithmic reasoning tasks and is publicly available at \url{https://github.com/deepmind/neural_networks_chomsky_hierarchy} under the Apache 2.0 License.
The tasks do not consist of fixed-size datasets but define training and testing distributions from which one can sample continuously.
We train the models for \num{2000000} steps with a batch size of 128, which corresponds to \num{256000000} (potentially non-unique) training examples.
At test time, we evaluate a single batch of size 500 for every sequence length in $\left\{41, \ldots, 500\right\}$, which corresponds to \num{230000} testing examples. 
We use the Adam optimizer~\citep{kingma2015adam} with gradient clipping and sweep over three learning rates: \num{1e-4}, \num{3e-4}, and \num{5e-4}.
Furthermore, for each task and positional encoding, we use 10 different parameter initialization random seeds.

We consider the encoder-only Transformer architecture~\citep{vaswani2017attention}, with 5 blocks of 8 heads each and $d_{\mathrm{model}} = 64$, which corresponds to \num{249026} parameters (\num{270146} in the case of relative and randomized relative positional encodings).
We run every task-encoding-hyperparameter triplet on a single NVIDIA V100 GPU from our internal cluster.
As a result, we used $15~(\mathrm{tasks}) \cdot 13~(\mathrm{positional~encodings}) \cdot 3~(\mathrm{learning~rates}) \cdot 10~(\mathrm{seeds}) = 5850$ GPU-units for the results in \cref{tab:scores,tab:scores-means-variances,tab:geometric,fig:scores-encodings}.
For the results in \cref{fig:efficiency}, we used an additional $2~(\mathrm{positional~encodings}) \cdot 3~(\mathrm{learning~rates}) \cdot 10~(\mathrm{seeds}) = 60$ GPU-units.
Finally, for \cref{fig:parameter}, we used $4~(\mathrm{maximum~positions}) \cdot 3~(\mathrm{learning~rates}) \cdot 10~(\mathrm{seeds}) = 120$ GPU-units, yielding a grand total of \num{6030} GPU-units.
We report all running times in \cref{tab:running-times} and observe that our method induces a negligible computational overhead.

\begin{table*}
    \caption{Mean and standard deviation of the running times (in hours) for all the positional encodings and tasks.}
    \begin{center}
        \resizebox{\linewidth}{!}{\input{tables/running_times}}
    \end{center}
    \label{tab:running-times}
\end{table*}

%% file: tables/running_times.tex
\begin{tabular}{@{}llcrrrrrrcrrrrr@{}}
    \toprule
    & & && & & & & && \multicolumn{5}{c}{\textbf{Randomized (Ours)}} \\
    \cmidrule{11-15}
    \textbf{Level} & \textbf{Task} && \textbf{None} & $\sin/\cos$ & \textbf{Relative} & \textbf{ALiBi} & \textbf{RoPE} & \textbf{Learned} && \textbf{$\sin/\cos$} & \textbf{Relative} & \textbf{ALiBi} & \textbf{RoPE} & \textbf{Learned$^\bigstar$} \\
    \midrule
    \multirow{4}{*}{R}

    & \paritycheck{}$^\dagger$ && $0.86\pm0.17$ & $0.87\pm0.17$ & $1.63\pm0.28$ & $0.87\pm0.17$ & $1.41\pm0.24$ & $0.90\pm0.18$ && $0.92\pm0.18$ & $1.75\pm0.29$ & $0.94\pm0.19$ & $1.66\pm0.31$ & $1.12\pm0.23$ \\
    & \reversestring{} && $1.17\pm0.21$ & $1.18\pm0.22$ & $2.61\pm0.39$ & $1.17\pm0.22$ & $2.01\pm0.35$ & $1.23\pm0.23$ && $1.24\pm0.23$ & $2.75\pm0.41$ & $1.27\pm0.24$ & $2.42\pm0.43$ & $1.62\pm0.32$ \\
    & \cyclenavigation{}$^\dagger$ && $0.86\pm0.17$ & $0.87\pm0.17$ & $1.62\pm0.27$ & $0.86\pm0.17$ & $1.41\pm0.25$ & $0.91\pm0.18$ && $0.92\pm0.18$ & $1.75\pm0.29$ & $0.94\pm0.19$ & $1.66\pm0.31$ & $1.12\pm0.22$ \\
    & \evenpairs{} && $0.86\pm0.17$ & $0.87\pm0.17$ & $1.63\pm0.27$ & $0.86\pm0.17$ & $1.41\pm0.24$ & $0.91\pm0.18$ && $0.92\pm0.18$ & $1.75\pm0.29$ & $0.95\pm0.19$ & $1.65\pm0.31$ & $1.12\pm0.22$ \\

    \\
    \multirow{4}{*}{DCF}
    & \stackmanipulation{} && $8.09\pm0.97$ & $8.00\pm0.82$ & $9.50\pm0.89$ & $8.07\pm0.94$ & $8.87\pm0.84$ & $8.46\pm0.84$ && $8.47\pm0.88$ & $10.04\pm0.96$ & $8.55\pm0.90$ & $10.61\pm1.58$ & $9.58\pm1.12$ \\
    & \modulararithmeticbrackets{} && $5.48\pm0.63$ & $5.55\pm0.67$ & $6.32\pm0.81$ & $5.50\pm0.65$ & $6.07\pm0.69$ & $5.69\pm0.65$ && $5.66\pm0.64$ & $6.56\pm0.70$ & $5.69\pm0.65$ & $6.41\pm0.84$ & $5.92\pm0.80$ \\
    & \binarymultiplication{} && $1.83\pm0.33$ & $1.83\pm0.30$ & $2.86\pm0.43$ & $1.84\pm0.31$ & $2.32\pm0.39$ & $2.24\pm0.35$ && $2.23\pm0.35$ & $3.13\pm0.43$ & $2.24\pm0.35$ & $3.21\pm0.51$ & $2.88\pm0.46$ \\
    & \binaryaddition{} && $1.83\pm0.32$ & $1.82\pm0.31$ & $2.89\pm0.42$ & $1.81\pm0.32$ & $2.34\pm0.39$ & $2.22\pm0.35$ && $2.22\pm0.35$ & $3.17\pm0.44$ & $2.24\pm0.35$ & $3.29\pm0.62$ & $2.90\pm0.49$ \\
    \\
    \multirow{7}{*}{CS}
    & \binaryaddition{} && $1.83\pm0.32$ & $1.82\pm0.31$ & $2.89\pm0.42$ & $1.81\pm0.32$ & $2.34\pm0.39$ & $2.22\pm0.35$ && $2.22\pm0.35$ & $3.17\pm0.44$ & $2.24\pm0.35$ & $3.29\pm0.62$ & $2.90\pm0.49$ \\
    & \computesqrt{} && $1.39\pm0.24$ & $1.40\pm0.25$ & $2.20\pm0.34$ & $1.40\pm0.25$ & $1.86\pm0.30$ & $1.73\pm0.29$ && $1.72\pm0.29$ & $2.43\pm0.37$ & $1.74\pm0.30$ & $2.53\pm0.41$ & $2.23\pm0.38$ \\
    & \solveequation{} && $5.60\pm0.65$ & $5.60\pm0.67$ & $6.41\pm0.68$ & $5.63\pm0.66$ & $6.14\pm0.68$ & $5.74\pm0.65$ && $5.78\pm0.66$ & $6.69\pm0.76$ & $5.83\pm0.69$ & $6.50\pm0.80$ & $6.01\pm0.84$ \\
    & \duplicatestring{} && $1.58\pm0.28$ & $1.59\pm0.28$ & $4.10\pm0.54$ & $1.58\pm0.27$ & $2.71\pm0.40$ & $1.64\pm0.28$ && $1.65\pm0.29$ & $4.24\pm0.54$ & $1.67\pm0.29$ & $3.18\pm0.49$ & $2.05\pm0.38$ \\
    & \modulararithmeticsimple{} && $0.99\pm0.19$ & $1.00\pm0.19$ & $1.74\pm0.29$ & $0.99\pm0.19$ & $1.51\pm0.26$ & $1.03\pm0.20$ && $1.05\pm0.20$ & $1.87\pm0.31$ & $1.06\pm0.21$ & $1.74\pm0.31$ & $1.23\pm0.23$ \\
    & \missingduplicate{} && $0.88\pm0.17$ & $0.90\pm0.18$ & $1.64\pm0.27$ & $0.88\pm0.17$ & $1.43\pm0.26$ & $0.93\pm0.19$ && $0.94\pm0.19$ & $1.78\pm0.30$ & $0.97\pm0.19$ & $1.66\pm0.30$ & $1.15\pm0.23$ \\
    & \oddsfirst{} && $1.17\pm0.22$ & $1.19\pm0.22$ & $2.61\pm0.38$ & $1.17\pm0.22$ & $2.00\pm0.31$ & $1.23\pm0.23$ && $1.24\pm0.23$ & $2.74\pm0.40$ & $1.26\pm0.23$ & $2.40\pm0.39$ & $1.59\pm0.29$ \\
    & \bucketsort{}$^\dagger$ && $1.17\pm0.23$ & $1.18\pm0.22$ & $2.61\pm0.43$ & $1.16\pm0.22$ & $2.01\pm0.34$ & $1.22\pm0.23$ && $1.24\pm0.23$ & $2.74\pm0.40$ & $1.25\pm0.23$ & $2.40\pm0.41$ & $1.60\pm0.30$ \\
    \bottomrule
\end{tabular}

%% file: 09_additional_results.tex
\section{Additional Results}
\label{sec:additional-results}

\subsection{Ablation Study}
\label{sec:additional-results:ablation}

In this section, we conduct an ablation study over the two main components of our method: (i) the maximum sampling position $L$, and (ii) the sorting of the subsampled positions.

\begin{figure}
    \begin{subfigure}[b]{0.5\textwidth}
        \begin{center}
            \includegraphics[width=\textwidth]{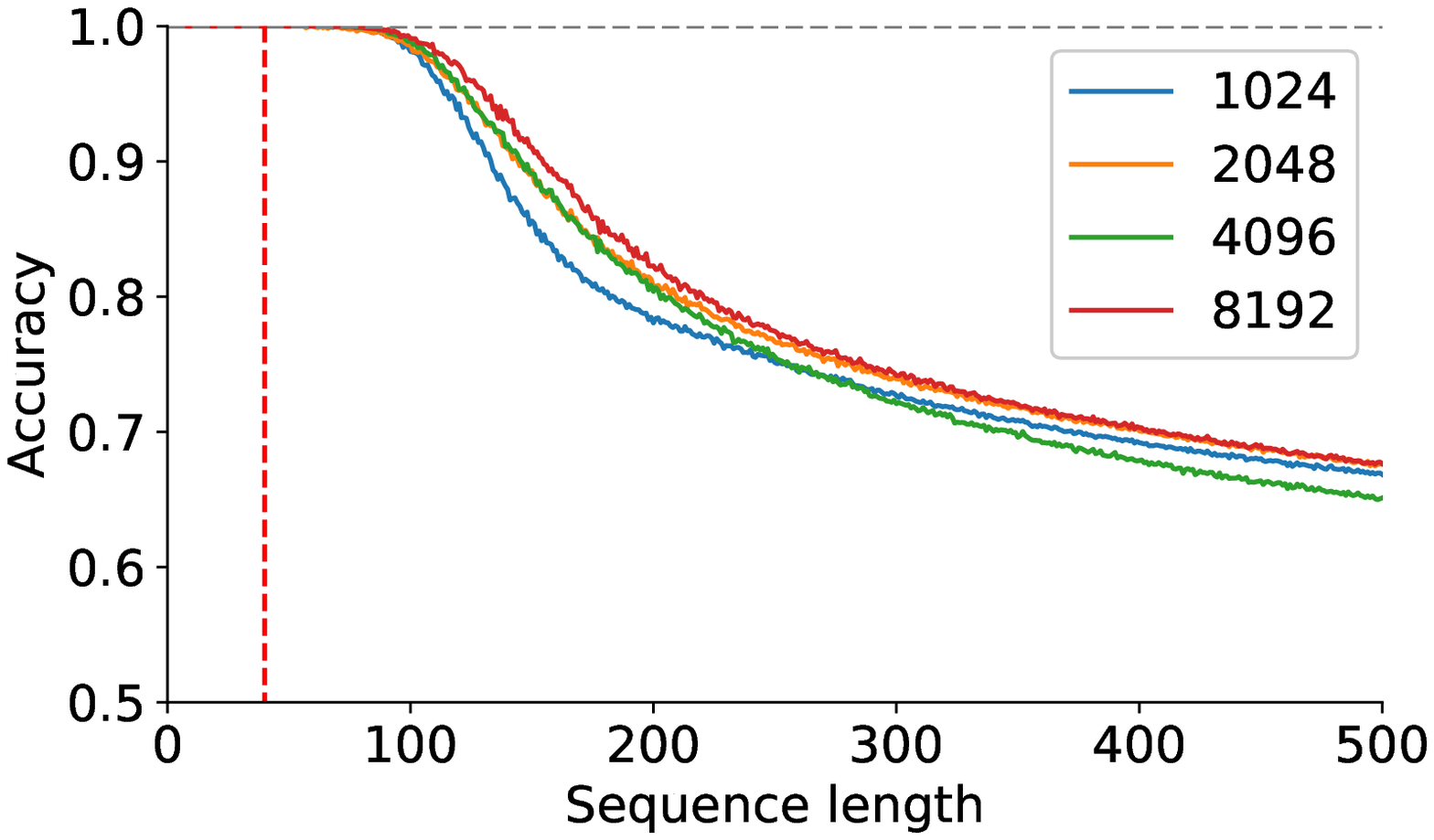}
        \end{center}
        \vspace{-0.25cm}
        \caption{\reversestring{}~(DCF)}
    \end{subfigure}
    
    \vspace{0.5cm}
    
    \begin{subfigure}[b]{0.5\textwidth}
        \begin{center}
            \includegraphics[width=\textwidth]{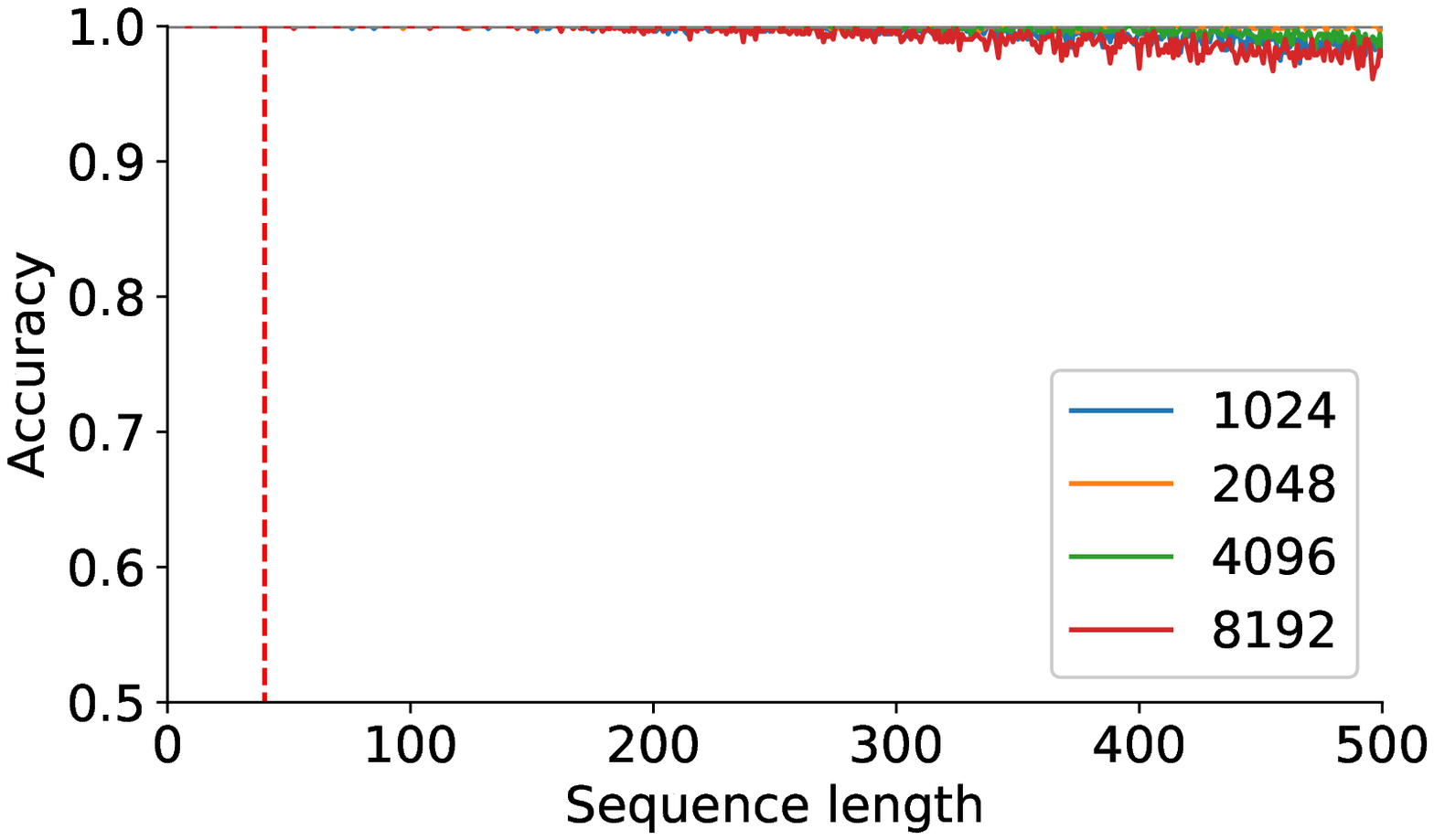}
        \end{center}
        \vspace{-0.25cm}
        \caption{\missingduplicate{}~(CS)}
    \end{subfigure}
    \caption{
        Sweep over the maximum position $L$ for our randomized relative positional encodings.
        The test accuracy (averaged over unseen sequence lengths) is largely unaffected by the concrete value of $L$ (for reasonably small values of $L$), showing the stability of our method.
        However, if $L$ is much larger than the maximum training ($N$) or testing ($M$) sequence length, we expect the performance to degrade since it the model is unlikely to encounter enough unique indices during training time. 
    }
    \label{fig:parameter}
\end{figure}

We train the randomized relative positional encoding for a wide range of different maximum positions $L$: \num{1024}, \num{2048}, \num{4096}, and \num{8192}.
\Cref{fig:parameter} shows that the test accuracy (averaged over all unseen sequence lengths) is largely unaffected by the value of $L$ on the \reversestring{} and \missingduplicate{} tasks.
As a consequence, a practitioner wanting to apply our method will not have to carry out extensive tuning of this parameter (as long as it is larger than the maximum evaluation sequence length $M$, but not unreasonably large).

Next, we investigate the performance of our randomized $\sin / \cos$ positional encoding with and without sorting of the subsampled positions.
Note that this experiment is meant as a ``sanity-check'' since we do not expect the Transformer to perform well without order information.
\cref{tab:ablation} shows the test accuracy (averaged over all unseen sequence lengths) for the two versions of our method.
We observe that sorting the positions is crucial, as it increases the test accuracy by 15.7\% on average and up to 76.3\% on certain tasks.
In fact, without sorting, our approach fails to beat the (baseline) random accuracy on all but the \cyclenavigation{} task, which is permutation-invariant (\ie it can be solved without positional information).
This confirms our intuition that the Transformer only needs to know the relative order of the positional encodings (and not their exact values), but that it fails to solve tasks when presented with positional encodings whose order does not correspond to the tokens' positions.

\begin{table}
    \caption{
        Accuracy (in percentage) averaged over all test lengths and maximized over 10 seeds and 3 learning rates for our randomized $\sin / \cos$ positional encoding with and without sorting of the subsampled positions.
    }
    \begin{center}
        \resizebox{\linewidth}{!}{\input{tables/ablation}}
    \end{center}
    \label{tab:ablation}
\end{table}

\subsection{Comparison to Prior Work}
\label{sec:additional-results:comparison}

In \cref{sec:experimental-evaluation}, we compared our method to a wide range of positional encodings: none, $\sin / \cos$~\citep{vaswani2017attention}, relative~\cite{dai2019transformer}, ALiBi~\cite{press2022train}, RoPE~\citep{su2021roformer}, learned~\citep{gehring2017convolutional}, and label-based~\citep{li2022systematic}.
Here, we provide additional results for these experiments, as well as a comparison to the geometric attention and directional encodings of \citet{csordas2022neural}.

\begin{figure*}
\begin{center}
        \begin{subfigure}[b]{0.31\textwidth}
            \begin{center}
                \includegraphics[width=\textwidth]{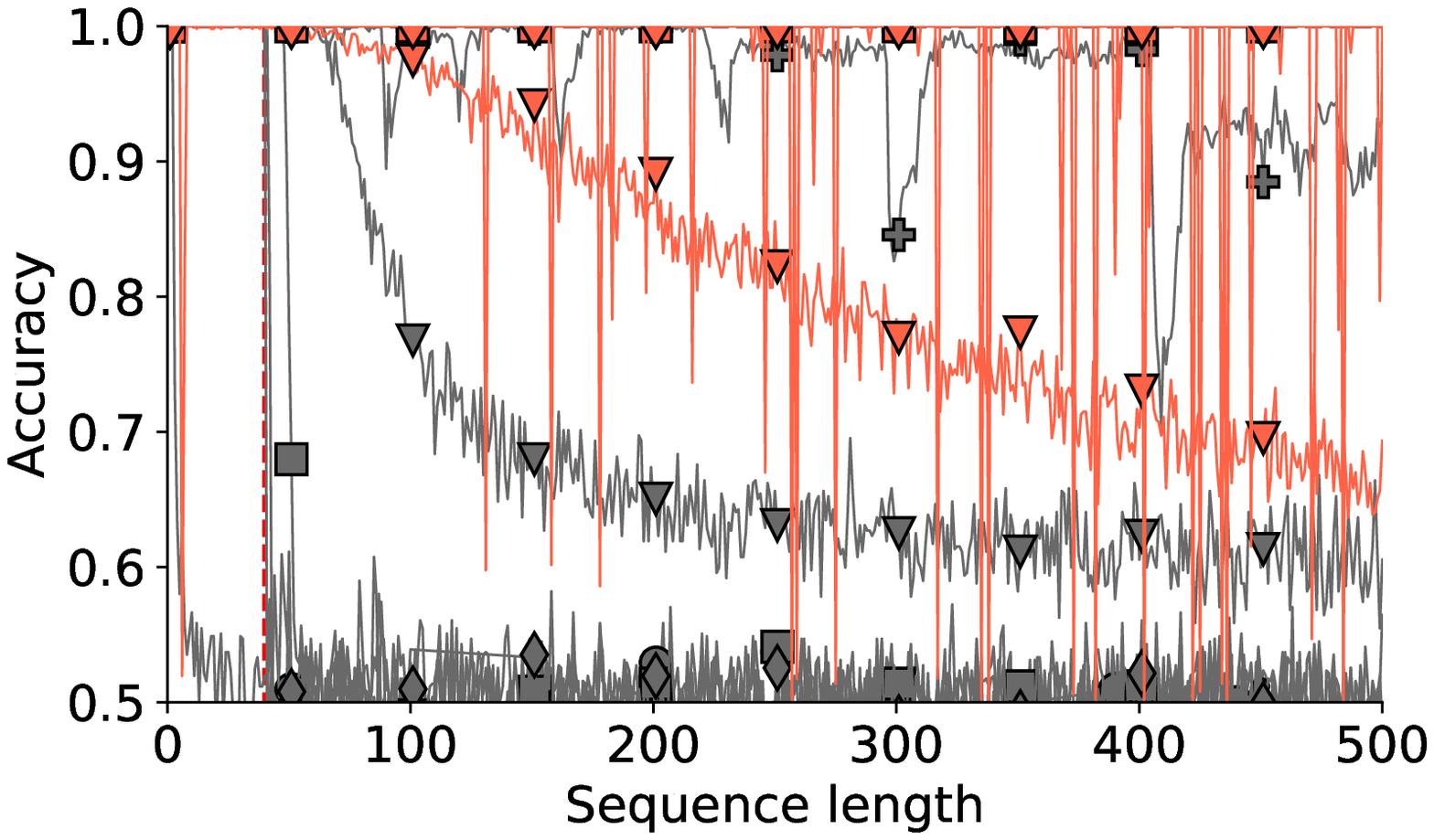}
            \end{center}
            \vspace{-0.25cm}
            \caption{\tiny{\evenpairs{} (R)}}
        \end{subfigure}
        \hspace{0.01\textwidth}
        \begin{subfigure}[b]{0.31\textwidth}
            \begin{center}
                \includegraphics[width=\textwidth]{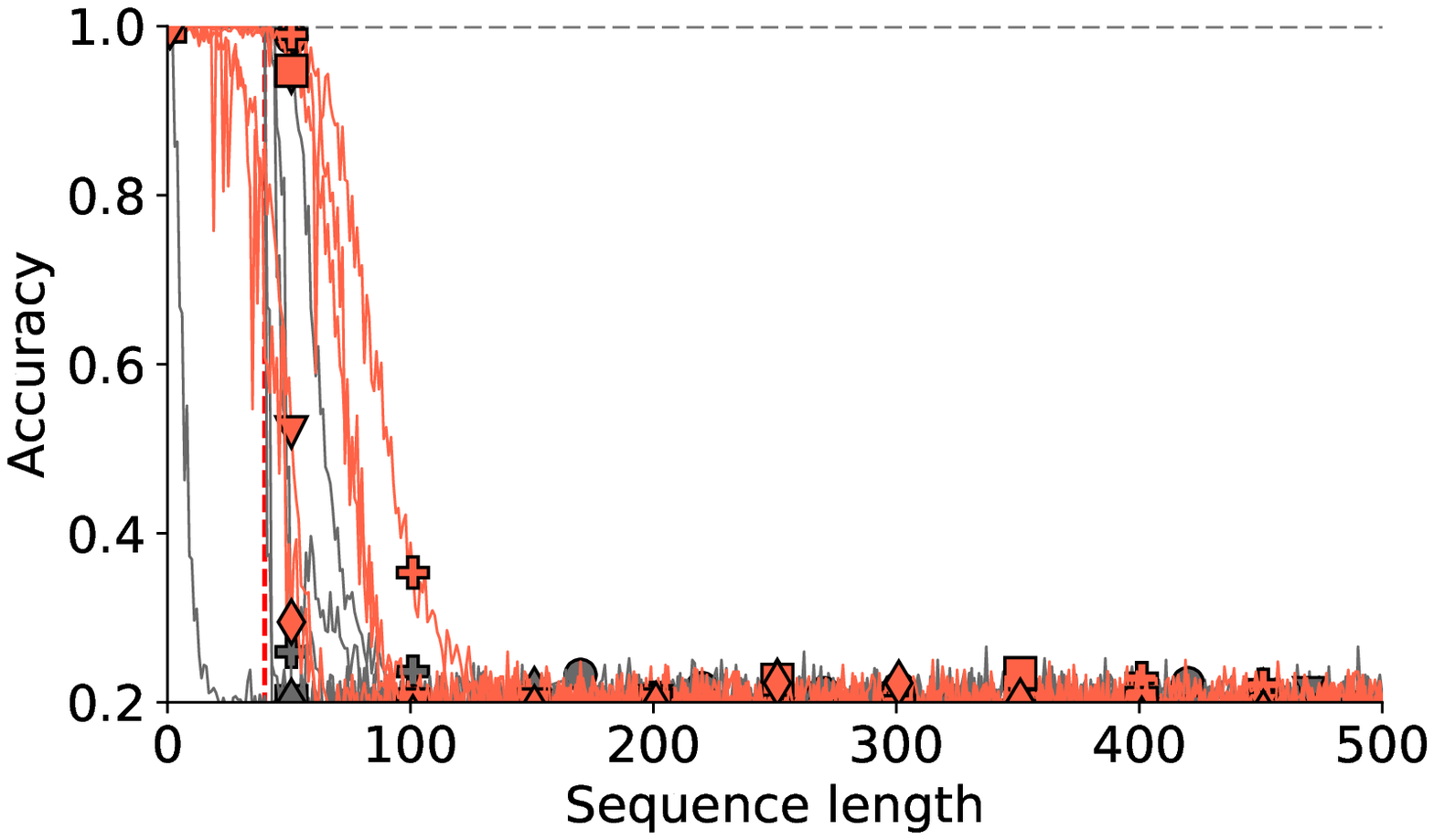}
            \end{center}
            \vspace{-0.25cm}
            \caption{\tiny{\modulararithmeticsimple{} (R)}}
        \end{subfigure}
        \hspace{0.01\textwidth}
        \begin{subfigure}[b]{0.31\textwidth}
            \begin{center}
                \includegraphics[width=\textwidth]{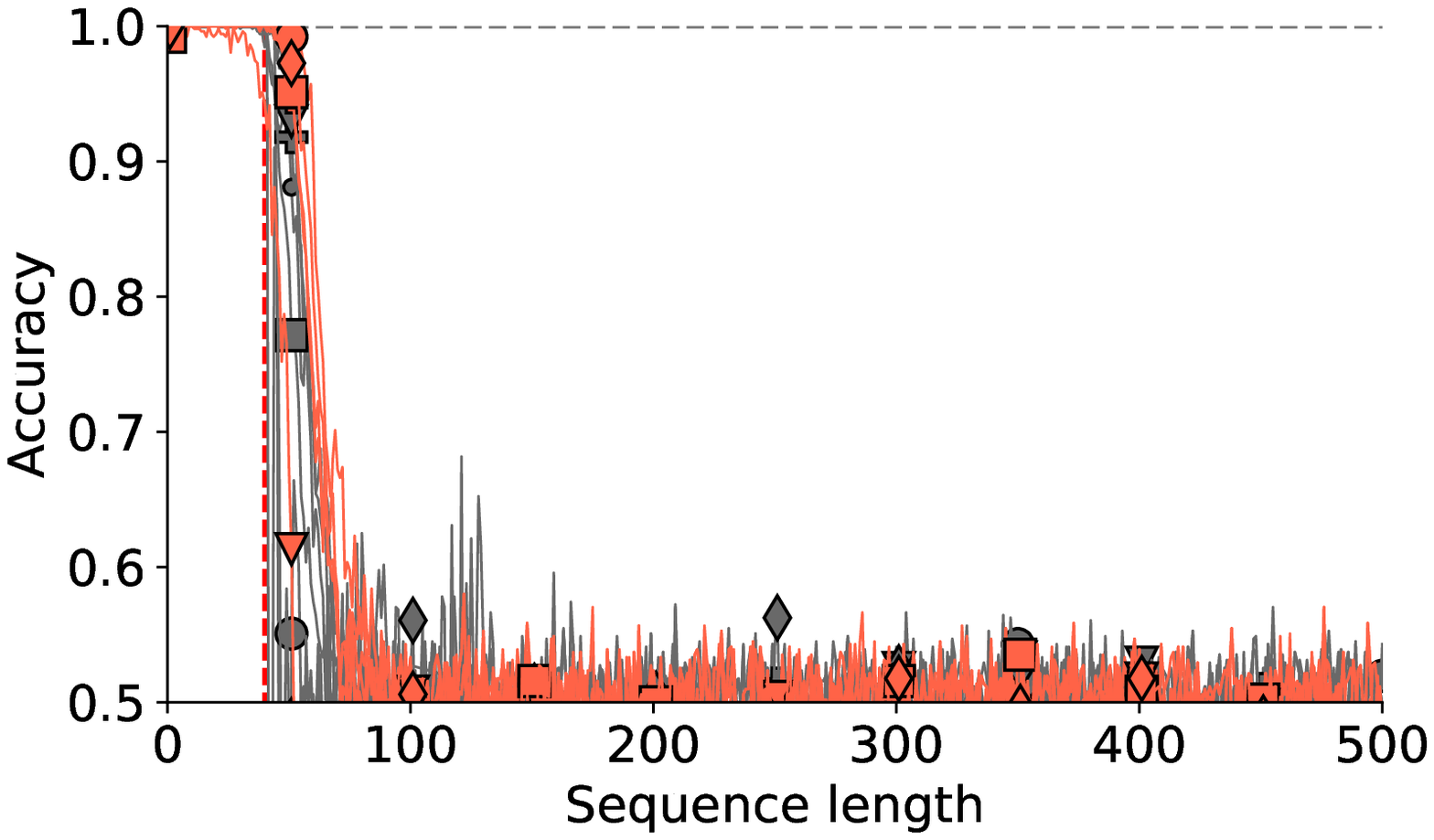}
            \end{center}
            \vspace{-0.25cm}
            \caption{\tiny{\paritycheck{} (R)}}
        \end{subfigure}
        \vspace{0.5cm}
        
        \begin{subfigure}[b]{0.31\textwidth}
            \begin{center}
                \includegraphics[width=\textwidth]{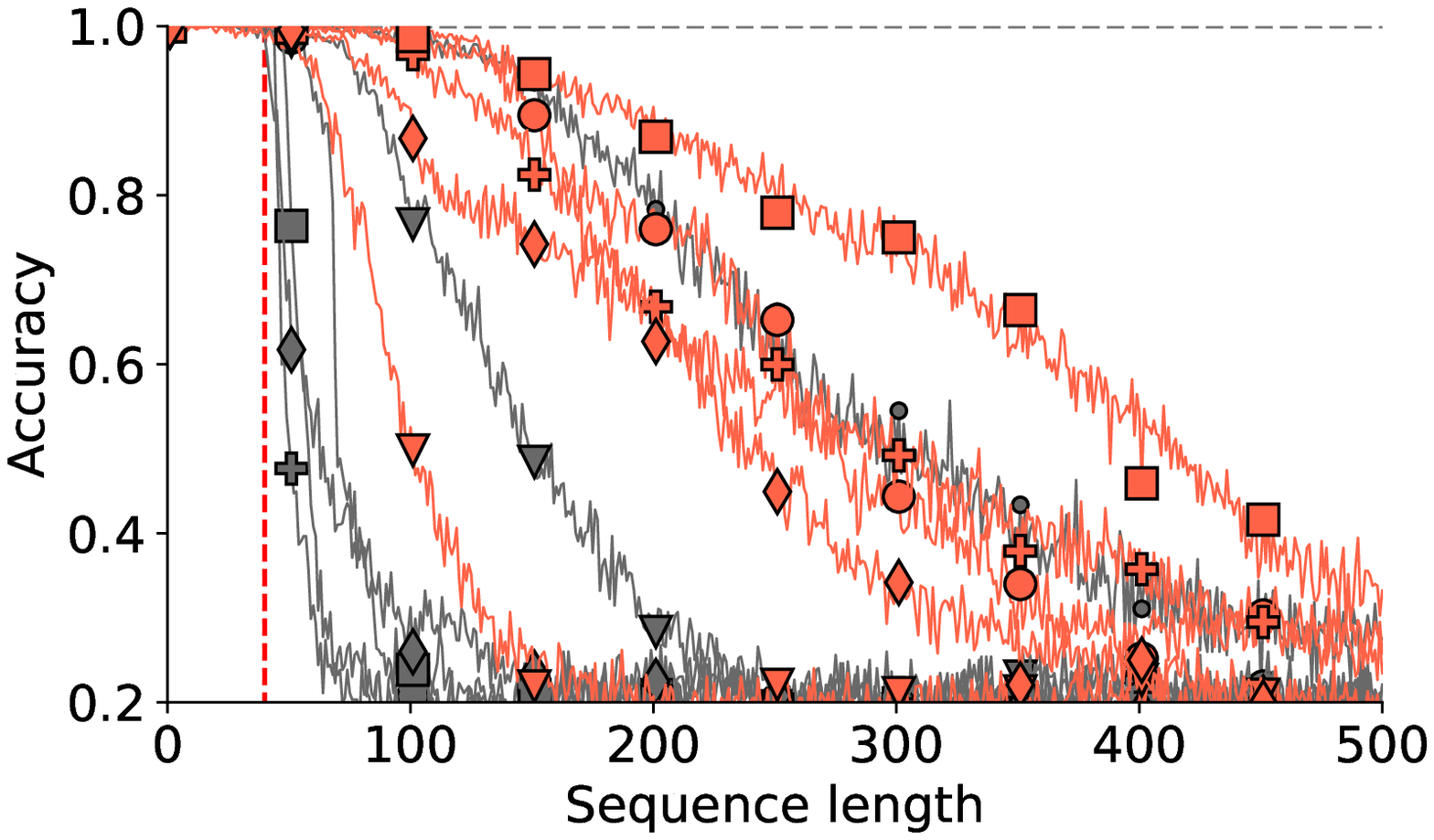}
            \end{center}
            \vspace{-0.25cm}
            \caption{\tiny{\cyclenavigation{} (R)}}
        \end{subfigure}
        \hspace{0.01\textwidth}
        \begin{subfigure}[b]{0.31\textwidth}
            \begin{center}
                \includegraphics[width=\textwidth]{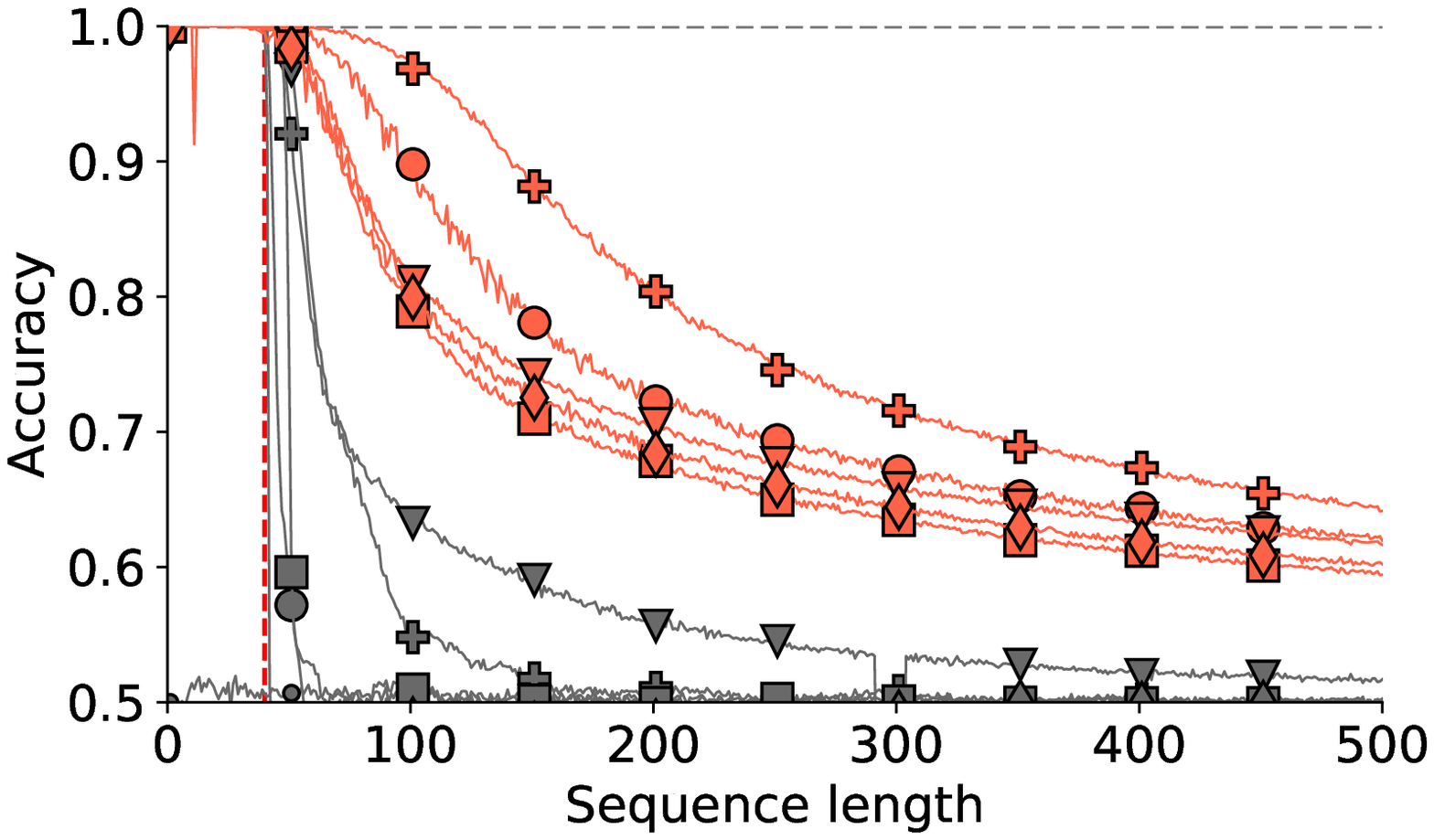}
            \end{center}
            \vspace{-0.25cm}
            \caption{\tiny{\stackmanipulation{} (DCF)}}
        \end{subfigure}
        \hspace{0.01\textwidth}
        \begin{subfigure}[b]{0.31\textwidth}
            \begin{center}
                \includegraphics[width=\textwidth]{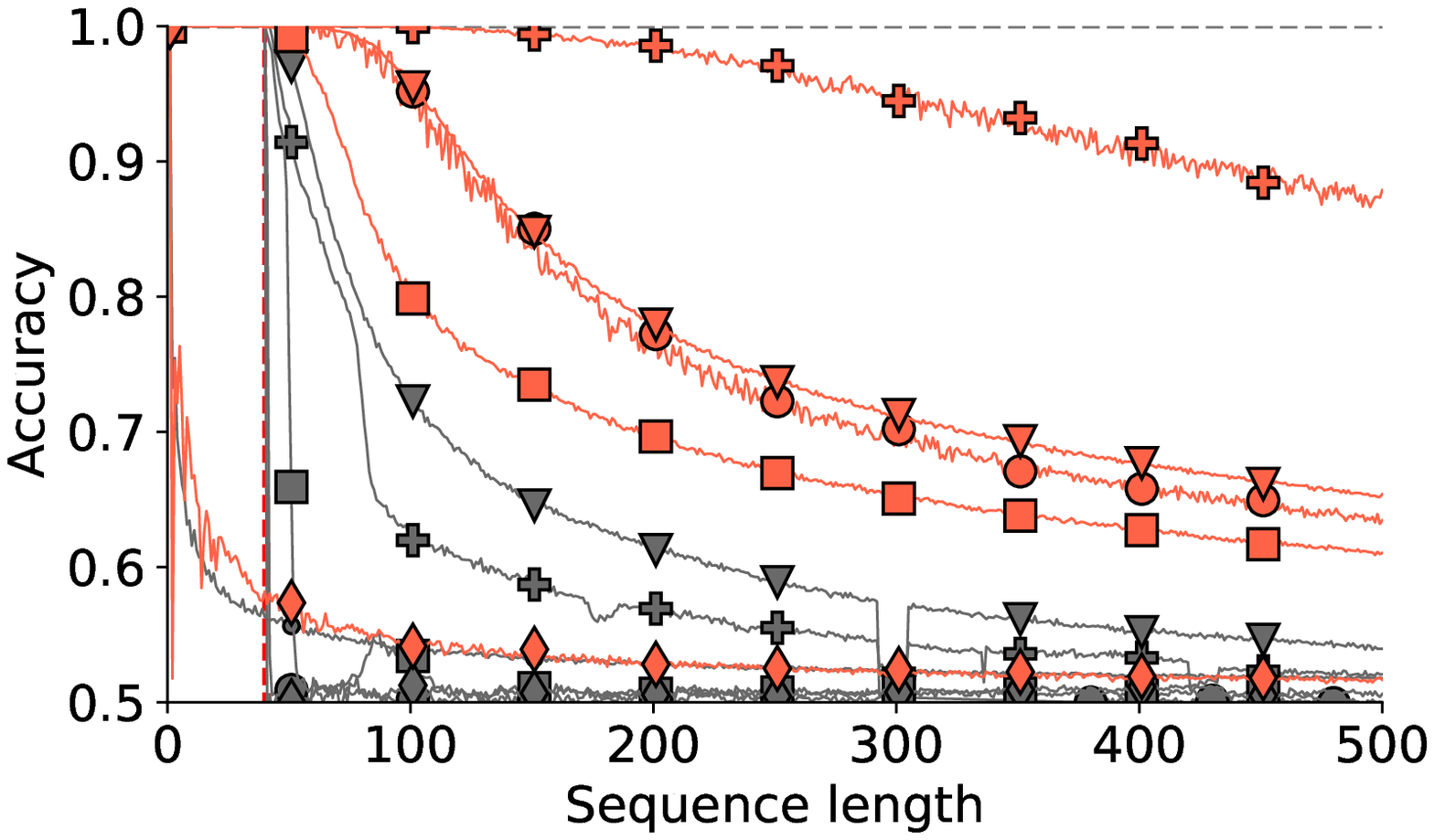}
            \end{center}
            \vspace{-0.25cm}
            \caption{\tiny{\reversestring{} (DCF)}}
        \end{subfigure}
        \vspace{0.5cm}
        
        \begin{subfigure}[b]{0.31\textwidth}
            \begin{center}
                \includegraphics[width=\textwidth]{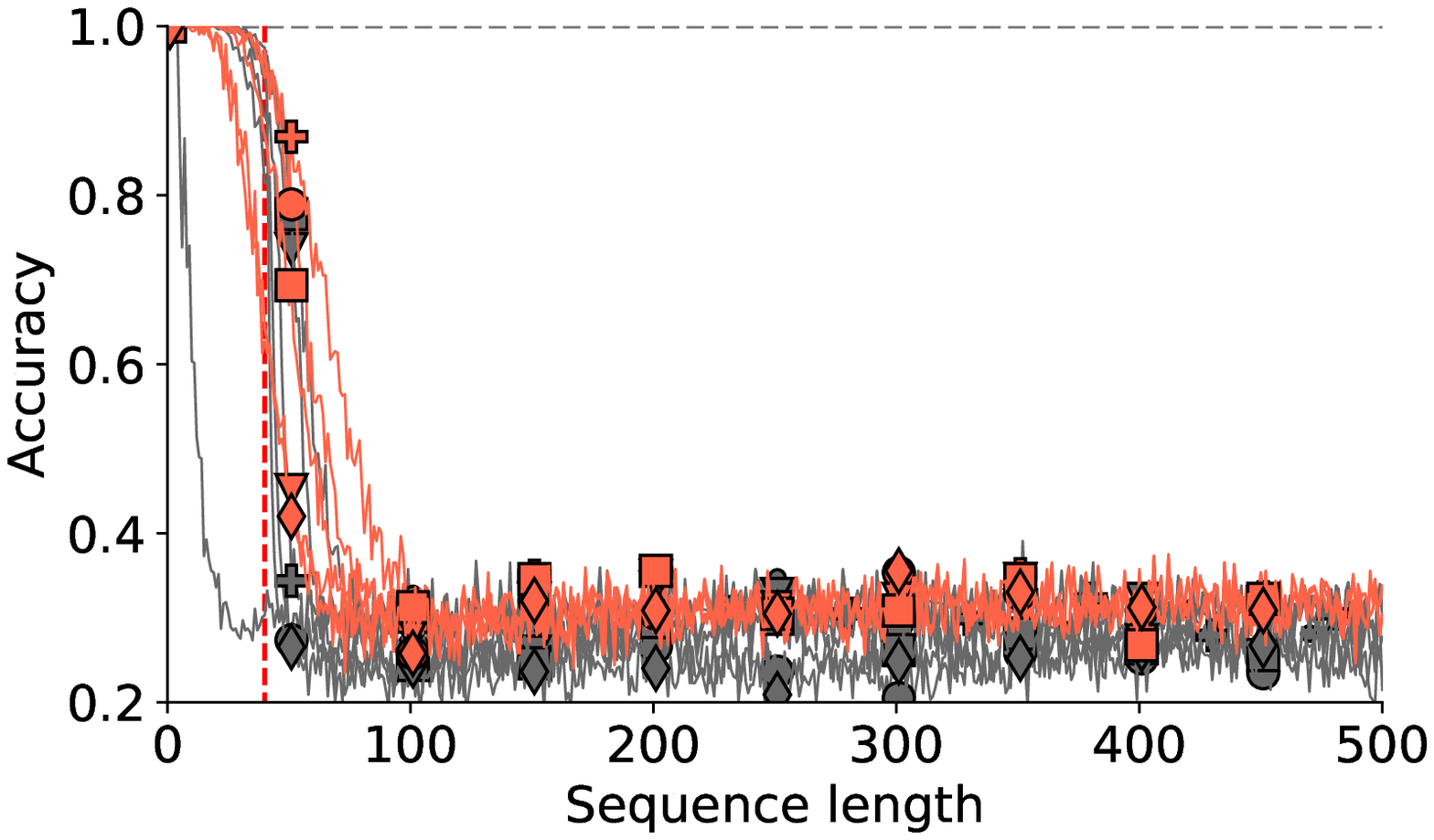}
            \end{center}
            \vspace{-0.25cm}
            \caption{\tiny{\modulararithmeticbrackets{} (DCF)}}
        \end{subfigure}
        \hspace{0.01\textwidth}
        \begin{subfigure}[b]{0.31\textwidth}
            \begin{center}
                \includegraphics[width=\textwidth]{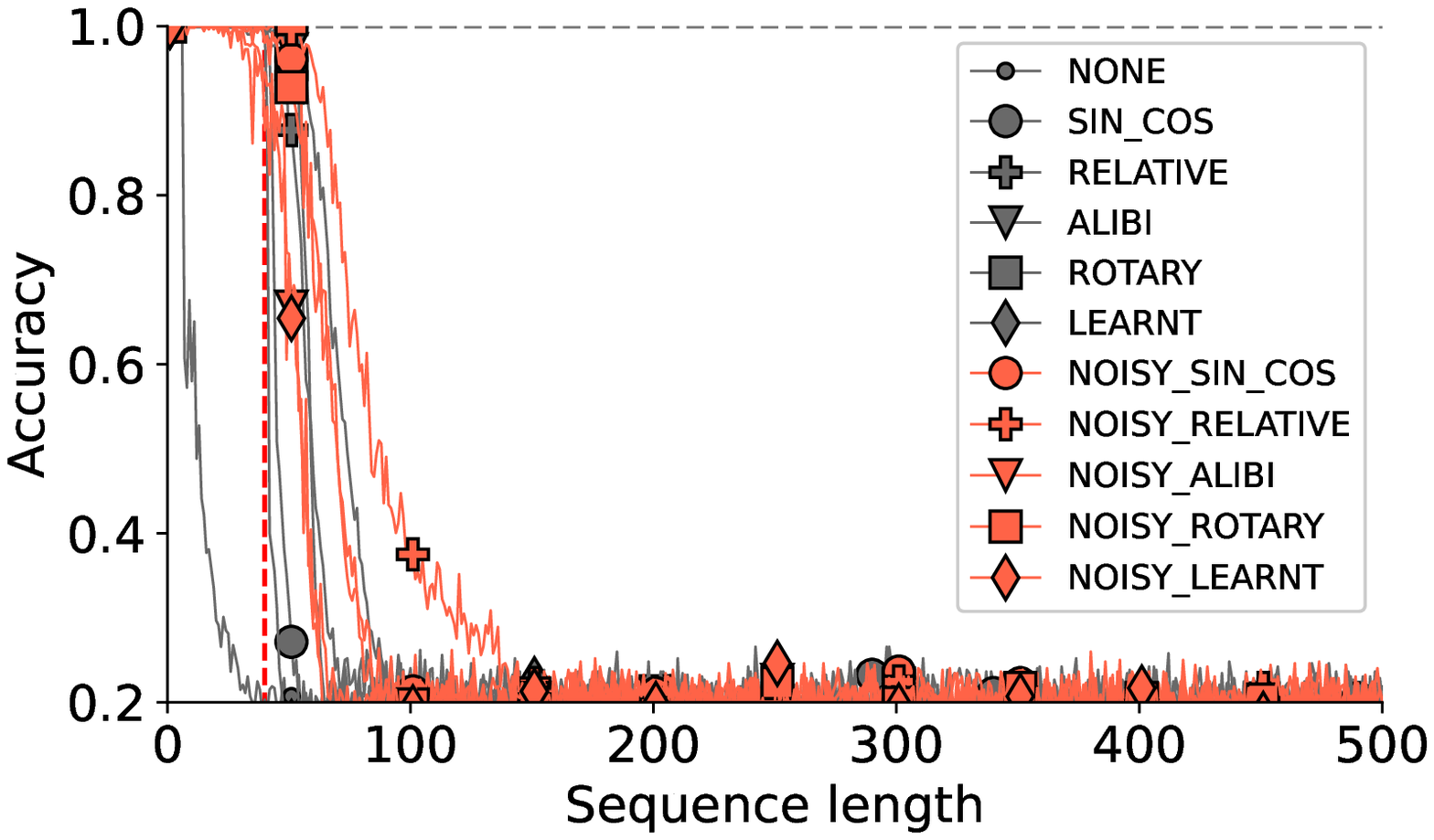}
            \end{center}
            \vspace{-0.25cm}
            \caption{\tiny{\solveequation{} (DCF)}}
        \end{subfigure}
        \hspace{0.01\textwidth}
        \begin{subfigure}[b]{0.31\textwidth}
            \begin{center}
                \includegraphics[width=\textwidth]{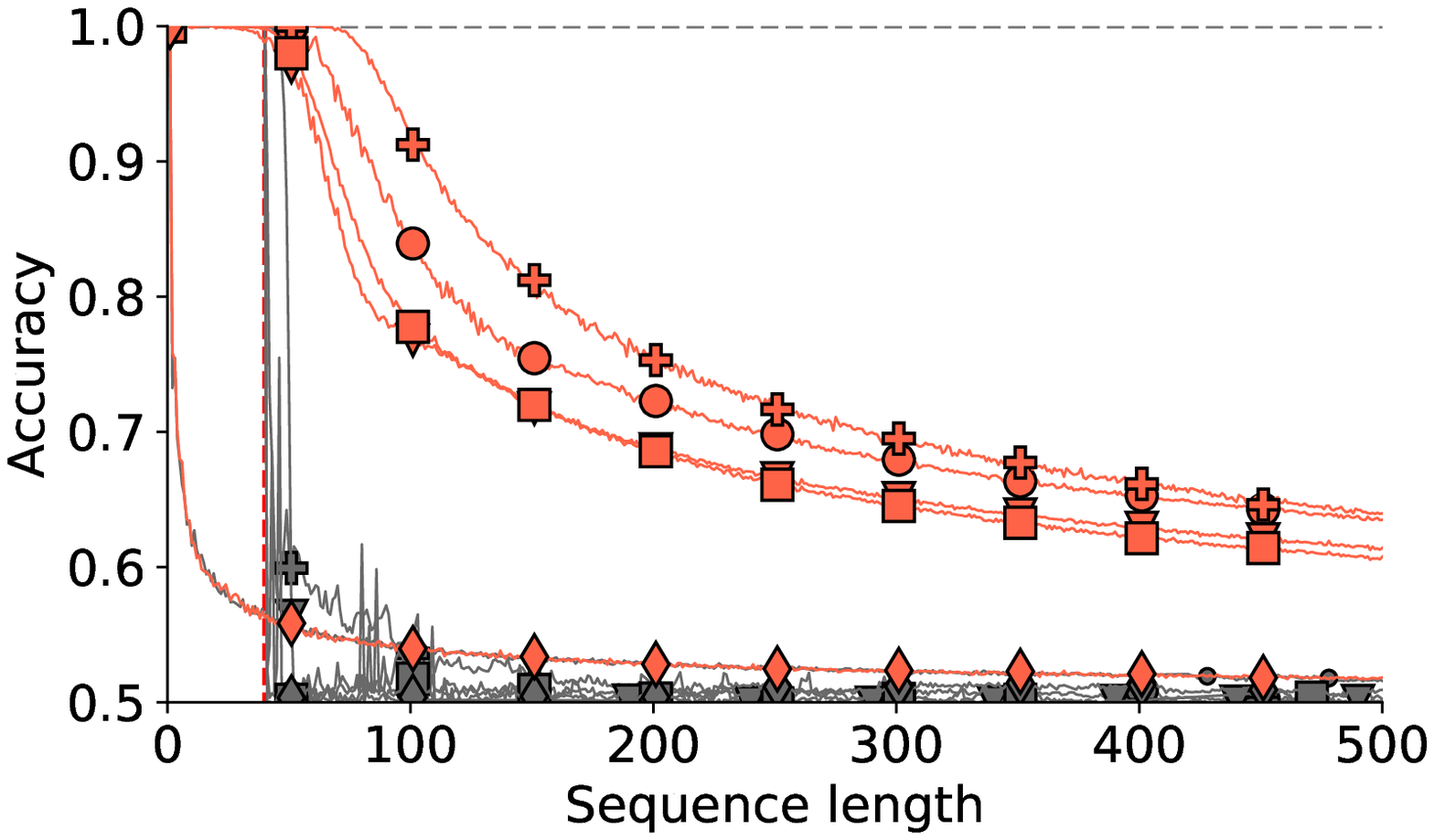}
            \end{center}
            \vspace{-0.25cm}
            \caption{\tiny{\duplicatestring{} (CS)}}
        \end{subfigure}
        \hspace{0.01\textwidth}
        \vspace{0.5cm}
        
        \begin{subfigure}[b]{0.31\textwidth}
            \begin{center}
                \includegraphics[width=\textwidth]{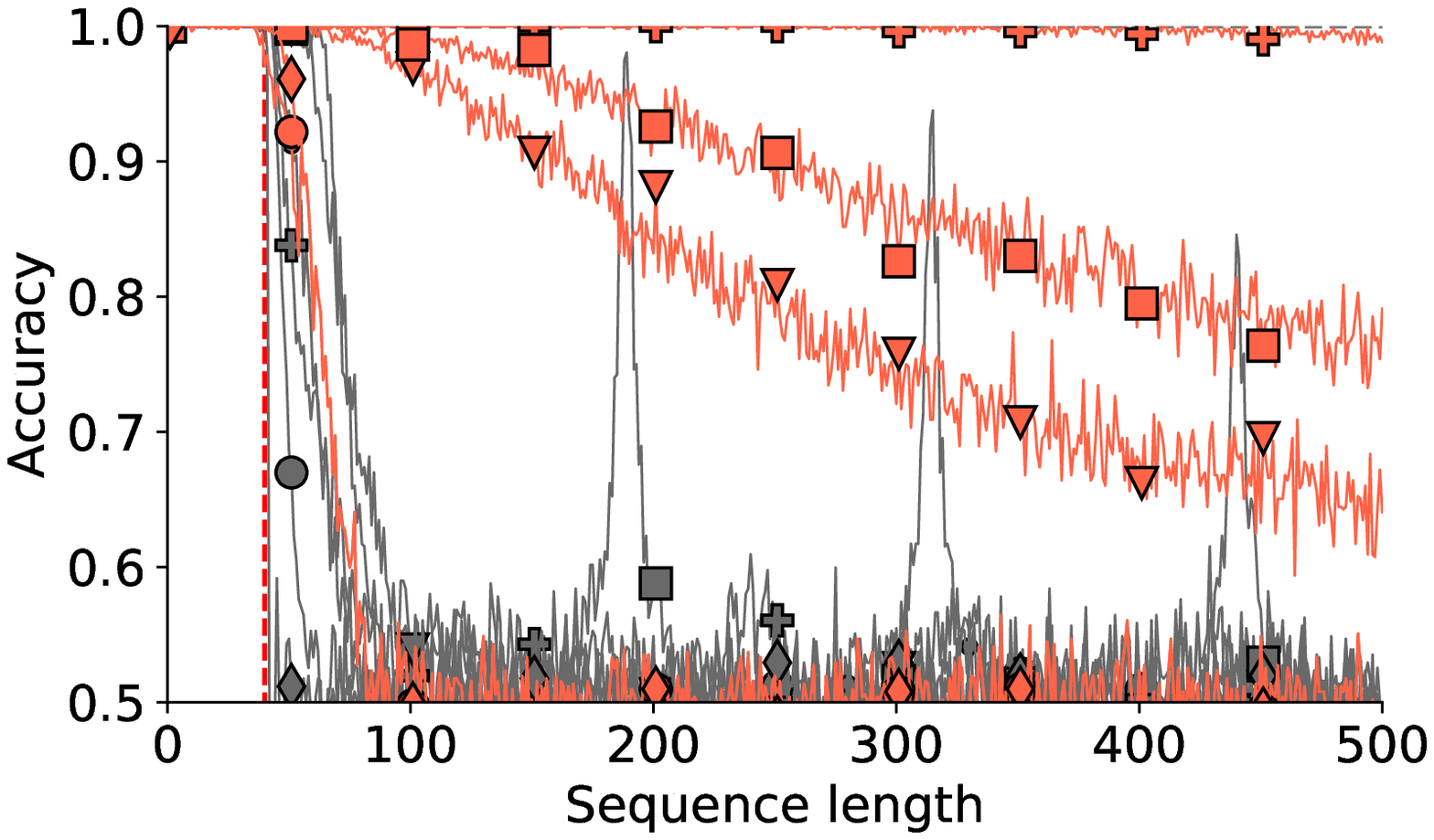}
            \end{center}
            \vspace{-0.25cm}
            \caption{\tiny{\missingduplicate{} (CS)}}
        \end{subfigure}
        \hspace{0.01\textwidth}
        \begin{subfigure}[b]{0.31\textwidth}
            \begin{center}
                \includegraphics[width=\textwidth]{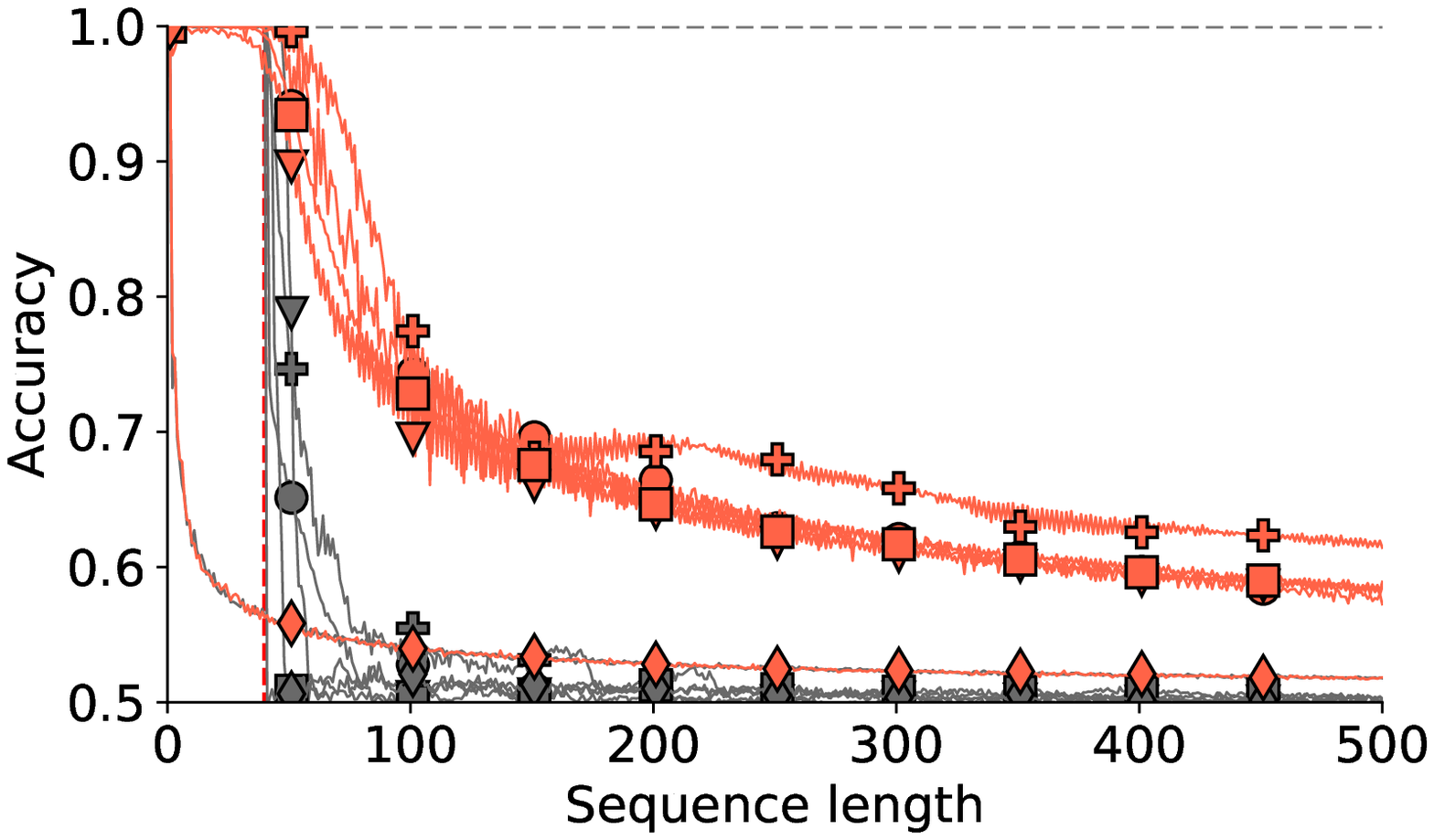}
            \end{center}
            \vspace{-0.25cm}
            \caption{\tiny{\oddsfirst{} (CS)}}
        \end{subfigure}
        \hspace{0.01\textwidth}
        \begin{subfigure}[b]{0.31\textwidth}
            \begin{center}
                \includegraphics[width=\textwidth]{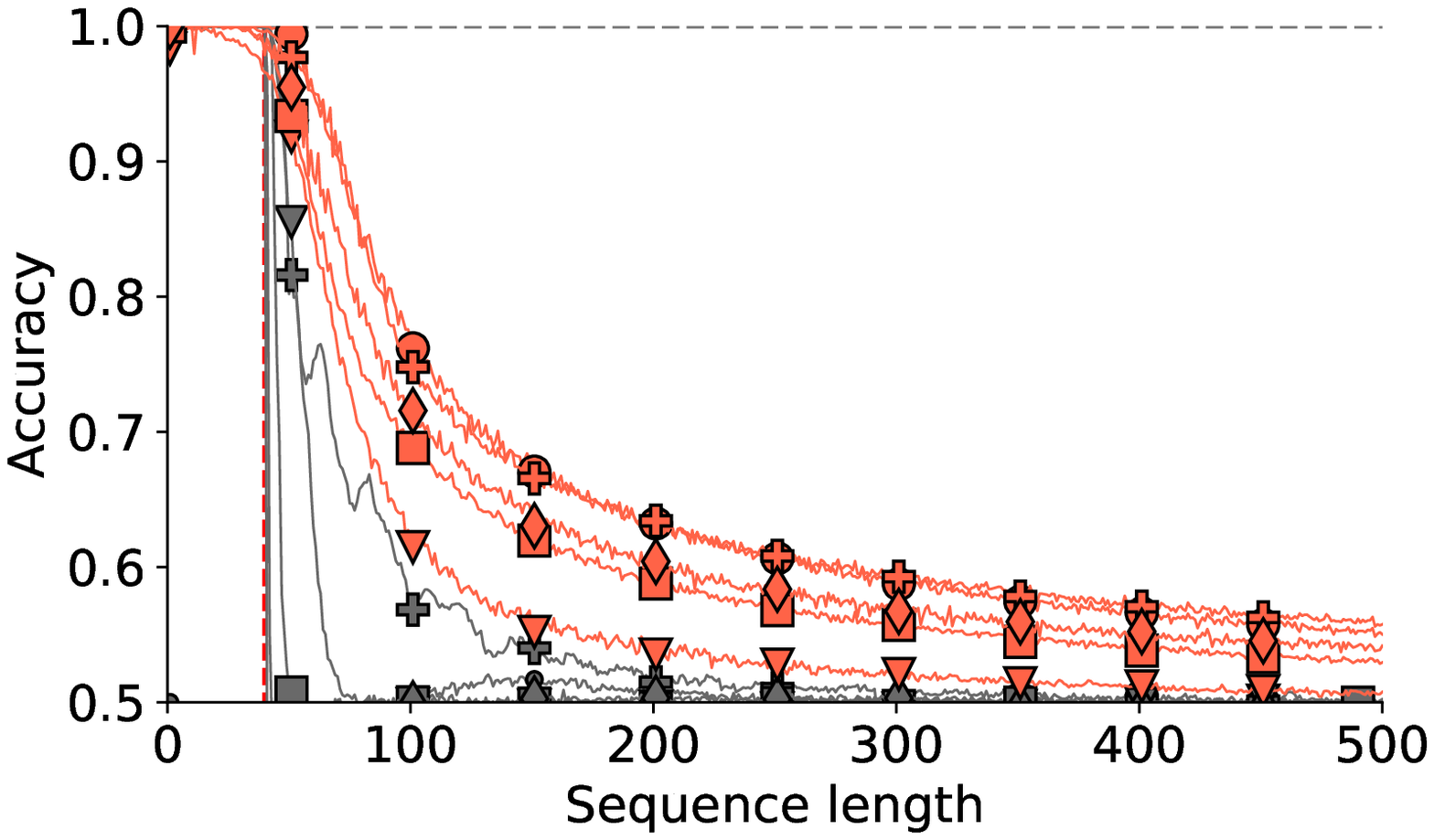}
            \end{center}
            \vspace{-0.25cm}
            \caption{\tiny{\binaryaddition{} (CS)}}
        \end{subfigure}
        \vspace{0.5cm}
        
        \begin{subfigure}[b]{0.31\textwidth}
            \begin{center}
                \includegraphics[width=\textwidth]{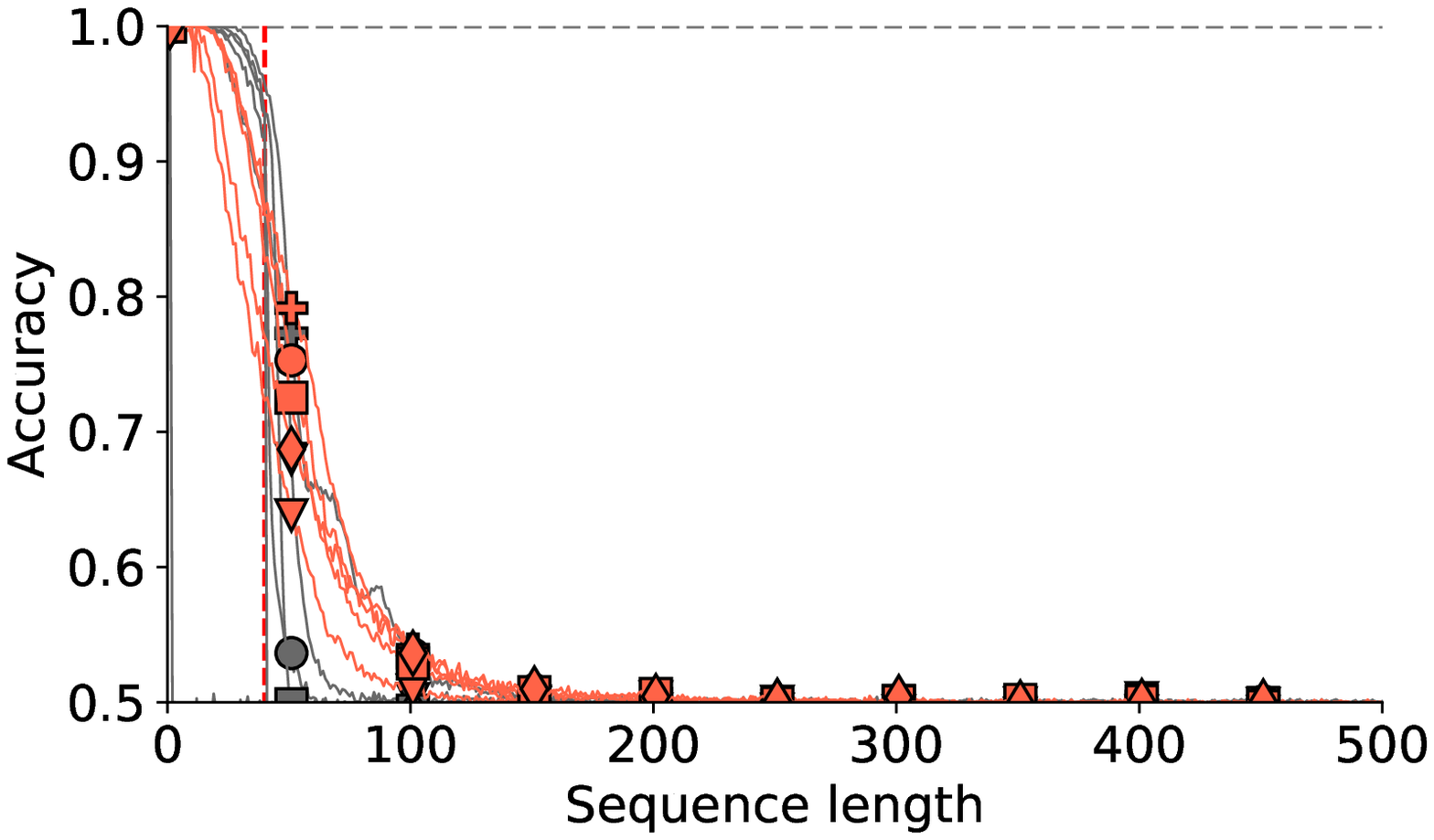}
            \end{center}
            \vspace{-0.25cm}
            \caption{\tiny{\binarymultiplication{} (CS)}}
        \end{subfigure}
        \hspace{0.01\textwidth}
        \begin{subfigure}[b]{0.31\textwidth}
            \begin{center}
                \includegraphics[width=\textwidth]{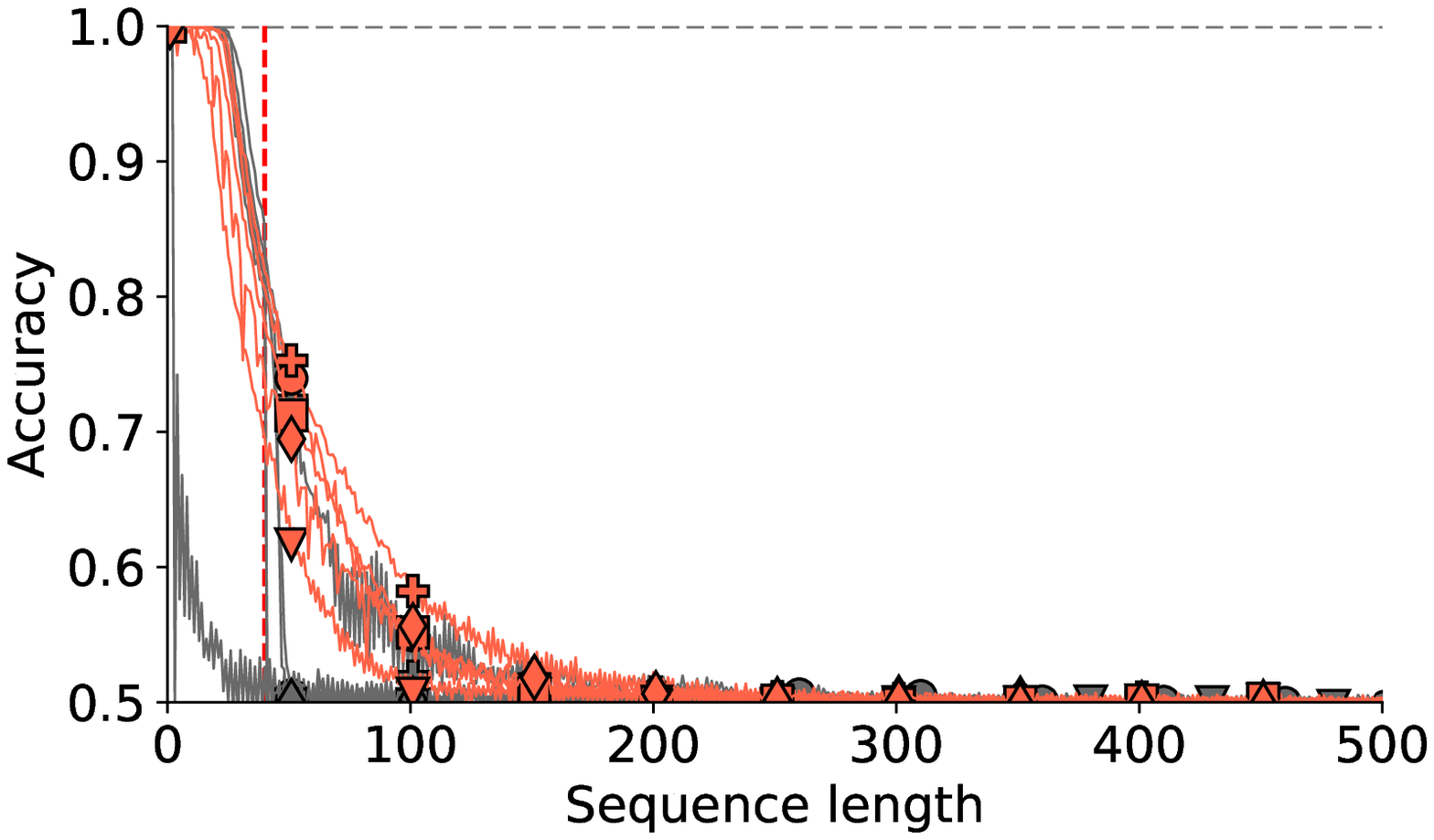}
            \end{center}
            \vspace{-0.25cm}
            \caption{\tiny{\computesqrt{} (CS)}}
        \end{subfigure}
        \hspace{0.01\textwidth}
        \begin{subfigure}[b]{0.31\textwidth}
            \begin{center}
                \includegraphics[width=\textwidth]{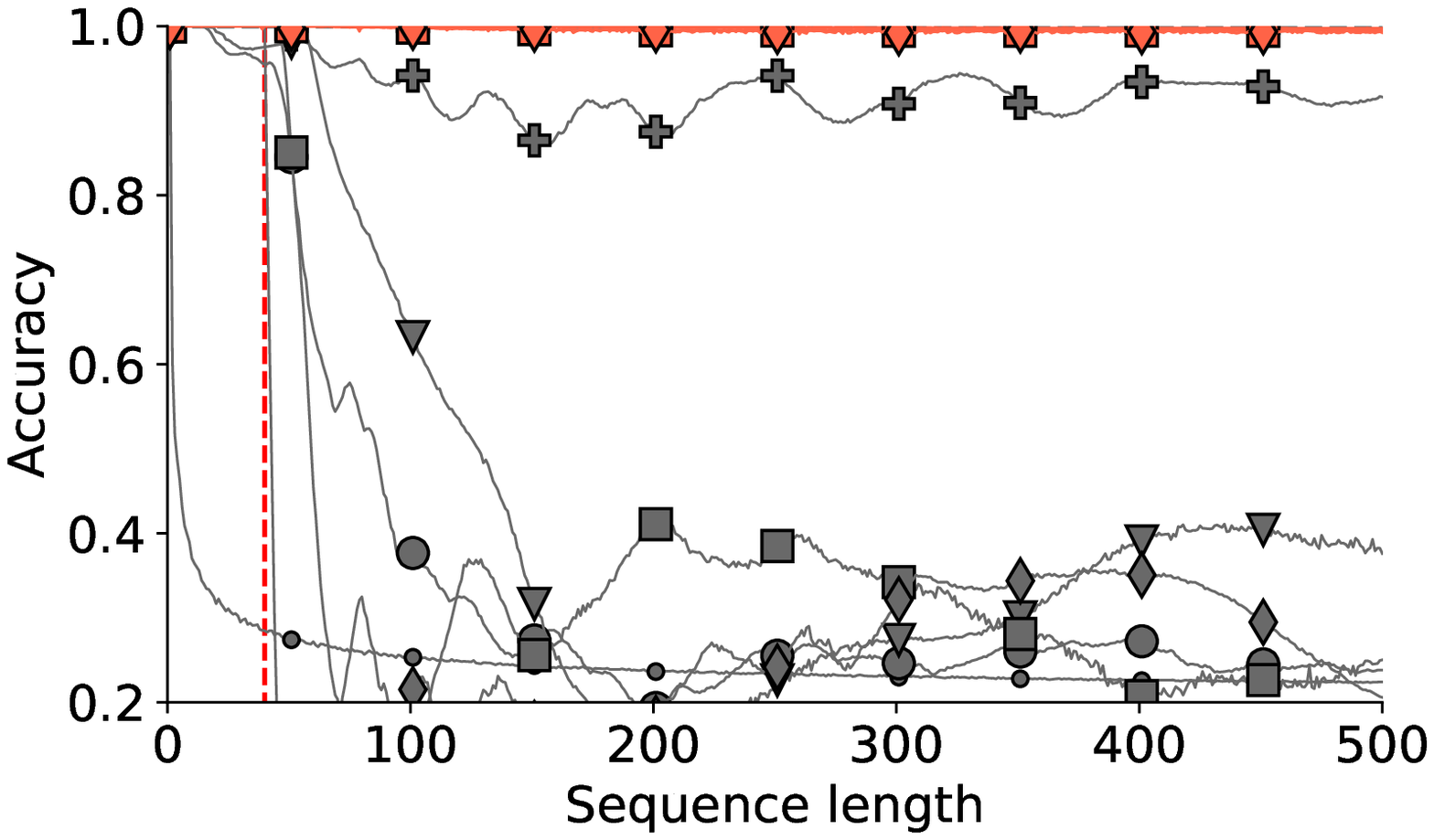}
            \end{center}
            \vspace{-0.25cm}
            \caption{\tiny{\bucketsort{} (CS)}}
        \end{subfigure}
        \hspace{0.01\textwidth}
    \end{center}
    \caption{
        Performance curves on all tasks for all the positional encodings.
        The dashed vertical red line is the training range, meaning that sequences to the right have not been seen during training and thus measure length generalization.
        The sequences to the left of the dashed line visualize the in-domain generalization performance.
    }
    \label{fig:scores-encodings}
\end{figure*}

We recall that \cref{tab:scores} showed the test accuracy maximized over the 10 parameter initialization seeds and the three different learning rates.
We reported the maximum following the experiment setup in \citet{deletang2023neural}, which investigates whether an architecture is capable of solving a task at all (and not on average).
However, we also report the means and standard deviations (over the random seeds) in \cref{tab:scores-means-variances} for the best-performing learning rate.
We observe that our randomized positional encoding also significantly outperform their original counterparts on average.
We visualize the test accuracy per sequence length in \cref{fig:scores-encodings}.

\begin{table*}
    \caption{
        Means and standard deviations (computed over random seeds) of the score (accuracy averaged over all test lengths) for the results of the main experiment (see \cref{tab:scores}).
        The random accuracy is 50\%, except for \cyclenavigation{}, \bucketsort{}, and the modular arithmetic tasks, where it is 20\%.
        We denote permutation-invariant tasks, which can be solved without positional information, with $\dagger$. Numbers in bold are the best performers, per task. These results underline the superiority of our method, and especially when applied to relative positional encodings.
    }
    \begin{center}
        \resizebox{\linewidth}{!}{\input{tables/scores_means_variances}}
    \end{center}
    \label{tab:scores-means-variances}
\end{table*}

We highlight the case of learned positional encodings, which fail to beat the random accuracy baseline (\cf \cref{tab:scores,tab:scores-means-variances}).
This is because the columns of the embedding matrix corresponding to the positions that are larger than the maximum training length $N$ are not learned during training and are thus entirely random.
In contrast, our randomized version of the learned encodings considers all possible embedding columns during training and thus achieves non-trivial to strong length generalization on most tasks.

Finally, we also compare our method to a variant of the Neural Data Router (NDR)~\citep{csordas2022neural}, which was developed to improve the systematic generalization capabilities of Transformers.
We only consider the most related aspects of the NDR architecture, \ie the geometric attention and the directional encoding (we do not use gating or shared layers).
\cref{tab:geometric} compares the test accuracy of geometric attention and directional encodings with the best results from \cref{tab:scores} (for the maximum) and \cref{tab:scores-means-variances} (for the mean).
We observe that our randomized positional encodings outperform the geometric attention overall (with a 9.7\% higher maximum test accuracy on average) but not on all tasks.
In particular, geometric attention performs substantially better on \modulararithmeticsimple{}, which has an inherent locality bias, \ie numbers closer to the operation symbols are generally more relevant, which can be captured by ``radiating outwards'' as geometric attention does.

\begin{table}
    \caption{
        Accuracy (in \%) averaged over all test lengths
        for geometric attention with directional encoding.
    }
    \begin{center}
        \resizebox{\linewidth}{!}{\input{tables/geometric}}
    \end{center}
    \label{tab:geometric}
\end{table}

\subsection{Analysis}
\label{sec:additional-results:analysis}

\paragraph{Analyzing the activations}

\begin{figure*}
    \begin{center}
        \begin{subfigure}[b]{\textwidth}
            \begin{center}
                \includegraphics[width=\textwidth]{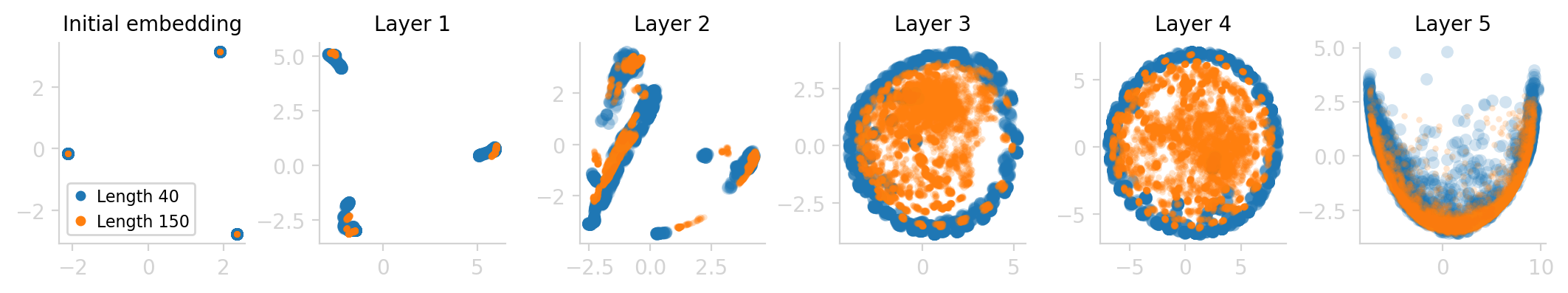}
            \end{center}
            \caption{Relative positional encoding~\citep{dai2019transformer}.}
        \end{subfigure}
        \par\bigskip
        \par\bigskip
        \begin{subfigure}[b]{\textwidth}
            \begin{center}
                \includegraphics[width=\textwidth]{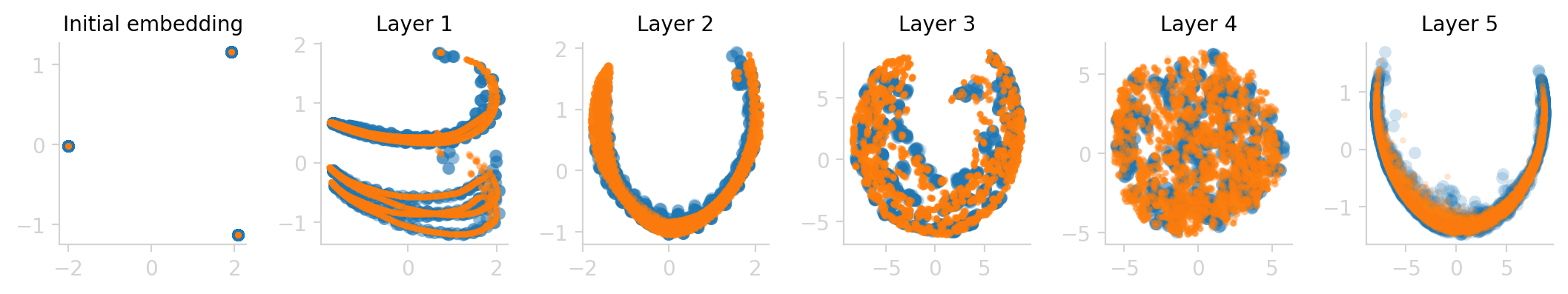}
            \end{center}
            \caption{Randomized relative positional encoding (ours).}
        \end{subfigure}
        \vspace{0.5cm}
    \end{center}
    \caption{
        2D PCA projections of the activations of the initial embeddings and the encoder layers for $30$~sequences on the \reversestring{} task.
        For sequence-lengths beyond the training length (shown in orange), the standard relative encoding clearly leads to out-of-distribution activations for layers 3 and 4 compared to those obtained with the maximum training length (shown in blue).
        In contrast, our randomized version does not lead to out-of-distribution activations for sequences longer than the maximum training length, confirming the intuition in~\cref{fig:overview}.
    }
    \label{fig:pca-encodings}
\end{figure*}

As illustrated in \cref{fig:overview}, the main intuition behind our randomized encodings is that they do not lead to out-of-distribution activations when evaluating on sequences longer than the maximal training length.
We confirm this intuition in our analysis in \cref{fig:pca-encodings}, which shows a 2D projection of activations onto the first two principal components when evaluating on sequences of length~$40$ (\ie the maximum training length $N$, shown in blue) and length~$150$ (\ie the generalization regime, shown in orange), using the same transformation.
While the activations of our randomized relative encoding strongly overlap for the training and the generalization regimes in all layers, the standard relative encoding leads to out-of-distribution activations for sequence length~$150$ in layers 3 and 4.
We obtained qualitatively similar results for the $\sin/\cos$ and learned encodings.

To compute the results in \cref{fig:pca-encodings}, we generated $30$ sequences of length $40$ and $150$ respectively, on the \reversestring{} task and passed them through a well-trained model with either relative or randomized relative encodings.
For each layer shown, we fitted a (non-whitened) 2D PCA on the activations obtained from sequence length~$40$ and projected all activations from sequence length~$150$ into two dimensions using the same transformations (yielding $30\times 40$ and $30\times 150$ activation-datapoints per layer).
The random relative encoding (our method) attains an average accuracy of $1.0$ and $0.994$ on the $30$ sequences of length $40$ and $150$, respectively.
The standard relative encoding (the baseline) attains an average accuracy of $1.0$ on sequence-length~$40$ and $0.596$ on length~$150$, indicating the model's failure to generalize well under the standard relative encoding.

\paragraph{Analyzing the attention matrices}

\begin{figure*}
\begin{center}
        \begin{subfigure}[b]{\textwidth}
            \begin{center}
                \includegraphics[width=\textwidth]{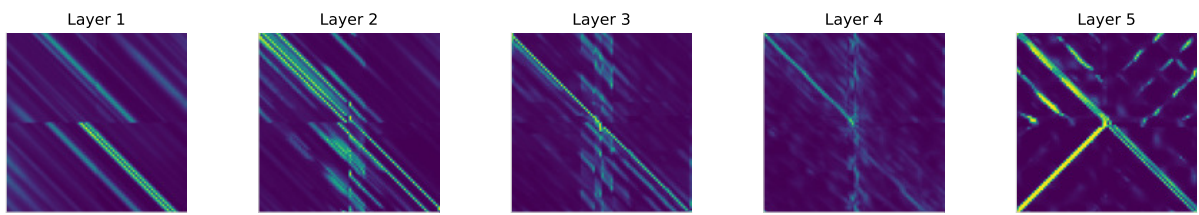}
            \end{center}
            \caption{Relative (baseline) with a sequence of length 40 (in-distribution).}
        \end{subfigure}
        \par\bigskip
        \par\bigskip
        \begin{subfigure}[b]{\textwidth}
            \begin{center}
                \includegraphics[width=\textwidth]{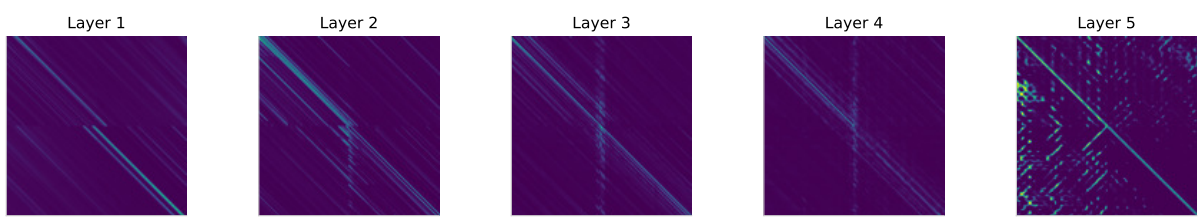}
            \end{center}
            \caption{Relative (baseline) with a sequence of length 150 (out-of-distribution).}
        \end{subfigure}
        \par\bigskip
        \par\bigskip
        \begin{subfigure}[b]{\textwidth}
            \begin{center}
                \includegraphics[width=\textwidth]{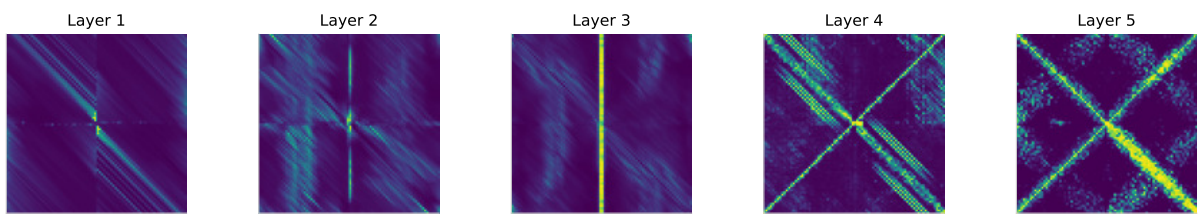}
            \end{center}
            \caption{Randomized relative (our method) with a sequence of length 40 (in-distribution).}
        \end{subfigure}
        \par\bigskip
        \par\bigskip
        \begin{subfigure}[b]{\textwidth}
            \begin{center}
                \includegraphics[width=\textwidth]{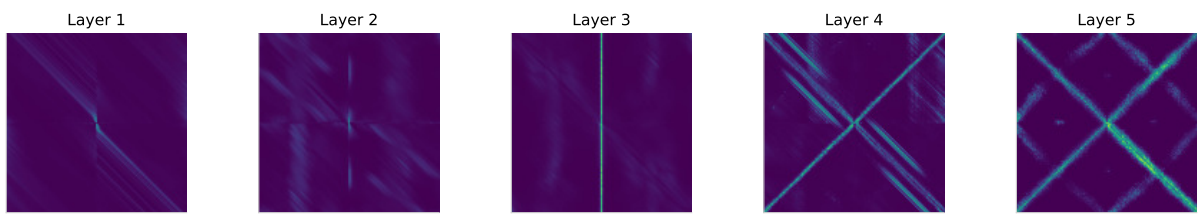}
            \end{center}
            \caption{Randomized relative (our method) with sequence of length 150 (out-of-distribution).}
        \end{subfigure}
    \end{center}
    \caption{
        Analysis of the attention matrices for the relative and randomized relative positional encodings on the \reversestring{} task using sequences of length 40 (\ie maximum training length) and 150 (\ie beyond training lengths).
        We visualize the maximum over the 8 heads per layer (following \citealp{csordas2022neural}) and observe a clear $X$ pattern, which corresponds to the reversal of the sequence.
        Our randomized relative encodings maintain that pattern on longer sequences, while it breaks down for the standard relative encoding.
    }
    \label{fig:attention-matrices}
\end{figure*}

We also analyze the attention matrices learned with the relative positional encoding and our corresponding randomized version on the \reversestring{} task.
To that end, we follow \citet{csordas2022neural} and visualize the maximum over the 8 attention matrices (one per head) for each of the 5 layers in \cref{fig:attention-matrices}.
We compare the attention matrices for sequences of length 40 (\ie the maximum training length) and 150 (\ie significantly longer than the maximum training length).
For length 40, both encodings produce a noticeable $X$ pattern, which corresponds to the reversal of the string.
However, for length 150, the pattern only remains visible for our randomized encodings while it breaks down for the original version, indicating the failure to generalize.

%% file: tables/ablation.tex
\begin{tabular}{@{}llcrr@{}}
    \toprule
    & && \multicolumn{2}{c}{\textbf{Randomized~$\sin / \cos$}} \\
    \cmidrule{4-5}
    \textbf{Level} & \textbf{Task} && \textbf{w/o Sorting} & \textbf{w/ Sorting} \\
    \midrule
    \multirow{4}{*}{R}
    & \evenpairs{}                  && 50.4 & \textbf{100.0} \\
    & \modulararithmeticsimple{}    && 20.0 & \textbf{25.7} \\
    & \paritycheck{}$^\dagger$      && 52.2 & \textbf{52.6} \\
    & \cyclenavigation{}$^\dagger$  && \textbf{59.3} & 59.0 \\
    \\
    \multirow{4}{*}{DCF}
    & \stackmanipulation{}          && 50.4 & \textbf{72.8} \\
    & \reversestring{}              && 52.8 & \textbf{75.6} \\
    & \modulararithmeticbrackets{}  && 31.0 & \textbf{33.8} \\
    & \solveequation{}              && 20.2 & \textbf{24.5} \\
    \\
    \multirow{7}{*}{CS}
    & \duplicatestring{}            && 52.8 & \textbf{72.4} \\
    & \missingduplicate{}           && \textbf{53.1} & 52.5 \\
    & \oddsfirst{}                  && 52.8 & \textbf{65.9} \\
    & \binaryaddition{}             && 50.0 & \textbf{64.4} \\
    & \binarymultiplication{}       && 49.9 & \textbf{52.1} \\
    & \computesqrt{}                && 50.2 & \textbf{52.5} \\
    & \bucketsort{}$^\dagger$       && 23.7 & \textbf{100.0} \\
    \bottomrule
\end{tabular}

%% file: tables/scores_means_variances.tex
\begin{tabular}{@{}llcrrrrrrcrrrrr@{}}
    \toprule
    & & && & & & & && \multicolumn{5}{c}{\textbf{Randomized (Ours)}} \\
    \cmidrule{11-15}
    \textbf{Level} & \textbf{Task} && \textbf{None} & $\sin/\cos$ & \textbf{Relative} & \textbf{ALiBi} & \textbf{RoPE} & \textbf{Learned} && \textbf{$\sin/\cos$} & \textbf{Relative} & \textbf{ALiBi} & \textbf{RoPE} & \textbf{Learned$^\bigstar$} \\
    \midrule
    \multirow{4}{*}{R}
 & \evenpairs{} && $50.1\pm0.1$  & $50.4\pm0.2$  & $67.6\pm15.3$  & $59.8\pm3.2$  & $50.4\pm0.3$  & $50.4\pm0.2$  && $99.7\pm0.3$  & $99.6\pm0.6$ & $71.4\pm5.6$ & $\textbf{100.0}\pm0.0$ & $96.2\pm0.7$  \\
 & \modulararithmeticsimple{} && $20.0\pm0.0$  & $20.2\pm0.2$  & $20.7\pm0.5$  & $23.2\pm0.9$  & $20.8\pm0.5$  & $20.1\pm0.1$  && $24.2\pm1.4$  & $\textbf{24.9}\pm1.7$ & $20.8\pm0.3$ & $23.5\pm1.6$ & $20.2\pm0.4$  \\
 & \paritycheck{}$^\dagger$ && $50.4\pm0.8$  & $50.3\pm0.2$  & $50.4\pm0.6$  & $50.5\pm0.6$  & $50.4\pm0.4$  & $50.0\pm0.1$  && $51.1\pm1.3$  & $\textbf{51.4}\pm0.5$ & $50.0\pm0.2$ & $50.4\pm1.0$ & $50.6\pm0.9$  \\
 & \cyclenavigation{}$^\dagger$ && $33.9\pm10.5$  & $23.8\pm1.4$  & $21.7\pm0.8$  & $31.1\pm3.8$  & $22.3\pm0.9$  & $21.0\pm1.2$  && $30.3\pm10.7$  & $45.9\pm9.9$ & $26.3\pm2.4$ & $\textbf{52.9}\pm15.3$ & $31.9\pm8.2$  \\
    \\
    \multirow{4}{*}{DCF}
 & \stackmanipulation{} && $50.2\pm0.1$  & $47.3\pm1.9$  & $50.1\pm3.3$  & $51.0\pm8.0$  & $49.6\pm3.0$  & $44.9\pm3.7$  && $69.2\pm3.2$  & $\textbf{71.7}\pm4.7$ &  $69.5\pm1.1$ & $66.0\pm2.0$ & $66.1\pm2.5$  \\
 & \reversestring{} && $52.7\pm0.1$  & $50.4\pm0.1$  & $54.2\pm1.5$  & $56.3\pm2.6$  & $51.2\pm0.3$  & $50.4\pm0.2$  && $72.9\pm1.6$  & $\textbf{77.1}\pm6.6$ & $75.1\pm1.3$ & $67.7\pm1.1$ & $52.7\pm0.2$  \\
 & \modulararithmeticbrackets{} && $31.0\pm0.1$  & $24.3\pm2.2$  & $26.1\pm2.0$  & $28.1\pm3.4$  & $24.0\pm2.4$  & $22.3\pm1.5$  && $29.6\pm4.6$  & $28.8\pm5.5$ & $29.3\pm1.6$ & $28.6\pm3.9$ & $\textbf{30.3}\pm2.6$  \\
 & \solveequation{} && $20.1\pm0.0$  & $20.9\pm0.2$  & $21.9\pm0.7$  & $23.6\pm1.9$  & $21.9\pm0.6$  & $20.2\pm0.2$  && $23.6\pm0.5$  & $\textbf{25.4}\pm1.8$ & $21.1\pm0.7$ & $22.3\pm1.6$ & $21.1\pm0.7$  \\
    \\
    \multirow{7}{*}{CS}
& \duplicatestring{} && $52.7\pm0.1$  & $50.4\pm0.2$  & $51.0\pm0.4$  & $51.0\pm0.2$  & $50.4\pm0.2$  & $50.4\pm0.2$  && $69.0\pm2.9$  & $\textbf{73.1}\pm1.5$ & $67.9\pm1.4$ & $67.1\pm2.0$ & $52.8\pm0.1$  \\
 & \missingduplicate{} && $51.4\pm1.0$  & $50.1\pm0.6$  & $51.1\pm1.1$  & $53.5\pm0.4$  & $53.9\pm1.6$  & $50.1\pm0.4$  && $50.4\pm1.5$  & $\textbf{91.4}\pm9.8$ &  $75.2\pm3.4$ & $73.2\pm1.2$ & $51.2\pm1.4$  \\
 & \oddsfirst{} && $52.7\pm0.1$  & $51.3\pm0.2$  & $51.5\pm0.5$  & $51.1\pm0.2$  & $50.8\pm0.2$  & $50.5\pm0.1$  && $62.5\pm2.0$  & $\textbf{65.9}\pm1.6$ & $62.2\pm1.4$ & $62.9\pm1.3$ & $52.7\pm0.1$  \\
 & \binaryaddition{} && $49.4\pm0.3$  & $47.3\pm3.8$  & $51.7\pm1.3$  & $48.5\pm3.6$  & $47.8\pm5.4$  & $48.9\pm0.8$  && $61.2\pm1.7$  & $\textbf{62.0}\pm1.1$ & $54.3\pm1.5$ & $57.4\pm1.2$ & $59.9\pm1.3$  \\
 & \binarymultiplication{} && $49.8\pm0.0$  & $48.8\pm1.0$  & $50.2\pm3.5$  & $49.9\pm2.3$  & $49.6\pm0.6$  & $48.7\pm1.7$  && $\textbf{51.8}\pm0.2$  & $39.1\pm7.1$ &  $49.2\pm1.2$ & $45.7\pm6.6$ & $51.6\pm0.2$  \\
 & \computesqrt{} && $50.2\pm0.0$  & $50.1\pm0.0$  & $51.5\pm0.4$  & $50.5\pm0.2$  & $50.3\pm0.1$  & $50.1\pm0.1$  && $51.9\pm0.5$  & $\textbf{52.4}\pm0.6$ &  $51.1\pm0.1$ & $51.8\pm0.3$ & $51.0\pm0.8$  \\
 & \bucketsort{}$^\dagger$ && $23.7\pm0.0$ & $25.6\pm2.6$ & $83.4\pm6.6$ & $29.3\pm6.7$ & $23.6\pm3.8$ & $20.7\pm2.9$  && $99.3\pm0.4$  & $\textbf{99.4}\pm0.3$ & $98.8\pm0.7$ & $99.3\pm0.3$ & $98.9\pm0.4$  \\
    \bottomrule
\end{tabular}

%% file: tables/geometric.tex
\begin{tabular}{@{}llcrrrr@{}}
    \toprule
    & && \multicolumn{2}{c}{\textbf{Max}} & \multicolumn{2}{c}{\textbf{Avg $\pm$ SD}} \\
    \cmidrule{4-5}
    \cmidrule{6-7}
    \textbf{Level} & \textbf{Task} && \textbf{\cref{tab:scores}} & \textbf{Geometric} & \textbf{\cref{tab:scores-means-variances}} & \textbf{Geometric} \\
    \midrule
    \multirow{4}{*}{R}
    & \evenpairs{}                  && \textbf{100.0} & \textbf{100.0} & $\bm{100.0}\pm0.0$ & $94.5\pm8.8$  \\
    & \modulararithmeticsimple{}    && 28.1           & \textbf{43.6}  & $24.9\pm1.7$       & $\bm{27.2}\pm8.2$  \\
    & \paritycheck{}$^\dagger$      && \textbf{52.6}  & 52.4           & $51.4\pm0.5$       & $\bm{51.6}\pm0.6$  \\
    & \cyclenavigation{}$^\dagger$  && \textbf{73.6}  & 41.3           & $\bm{52.9}\pm15.3$ & $32.9\pm4.7$  \\
    \\
    \multirow{4}{*}{DCF}
    & \stackmanipulation{}          && \textbf{77.9}  & 58.3           & $\bm{71.7}\pm4.7$  & $55.6\pm2.3$  \\
    & \reversestring{}              && \textbf{95.1}  & 65.2           & $\bm{77.1}\pm6.6$  & $59.3\pm3.2$  \\
    & \modulararithmeticbrackets{}  && 34.9           & \textbf{36.5}  & $30.3\pm2.6$       & $\bm{32.8}\pm2.8$  \\
    & \solveequation{}              && 28.1           & \textbf{31.7}  & $25.4\pm1.8$       & $\bm{28.5}\pm2.0$  \\
    \\
    \multirow{7}{*}{CS}
    & \duplicatestring{}            && \textbf{75.1}  & 58.6           & $\bm{73.1}\pm1.5$  & $54.9\pm1.6$  \\
    & \missingduplicate{}           && \textbf{100.0} & 64.4           & $\bm{91.4}\pm9.8$  & $60.3\pm2.3$  \\
    & \oddsfirst{}                  && \textbf{69.3}  & 64.2           & $\bm{65.9}\pm1.6$  & $58.1\pm2.6$  \\
    & \binaryaddition{}             && \textbf{64.5}  & 54.9           & $\bm{62.0}\pm1.1$  & $53.5\pm1.5$  \\
    & \binarymultiplication{}       && 50.1           & \textbf{53.6}  & $51.8\pm0.2$       & $\bm{52.1}\pm2.5$  \\
    & \computesqrt{}                && 53.3           & \textbf{54.1}  & $\bm{52.4}\pm0.6$  & $52.3\pm0.9$  \\
    & \bucketsort{}$^\dagger$       && \textbf{100.0} & 78.3           & $\bm{99.5}\pm0.3$  & $57.7\pm11.4$ \\
    \bottomrule
\end{tabular}